\documentclass[lettersize,journal]{IEEEtran}
\usepackage{amsmath,amsfonts}
\usepackage{algorithm}
\usepackage{algpseudocode} % 使用 algpseudocode 包
\usepackage{array}
\usepackage[caption=false,font=normalsize,labelfont=sf,textfont=sf]{subfig}
\usepackage{textcomp}
\usepackage{stfloats}
\usepackage{url}
\usepackage{verbatim}
\usepackage{graphicx}
\usepackage{cite}
\usepackage{microtype}
\usepackage{booktabs}
\usepackage{arydshln}
\usepackage{multirow}
\usepackage{xcolor} % 提供颜色支持，并加载表格颜色功能
\begin{document}

\title{Q-SiT: Teaching LMMs for Image Quality \\  Scoring and Interpreting}

\author{Zicheng Zhang, Haoning Wu, Ziheng Jia,\\ 
Weisi Lin,~\IEEEmembership{Fellow,~IEEE,} and Guangtao Zhai$^\dagger$,~\IEEEmembership{Fellow,~IEEE}
\IEEEcompsocitemizethanks{\IEEEcompsocthanksitem Zicheng Zhang, Ziheng Jia, and Guangtao Zhai are with the Institute of Image Communication and Network Engineering, Shanghai Jiao Tong University, China. E-mail:\{zzc1998,jzhws1,zhaiguangtao\}
@sjtu.edu.cn. \protect}
\IEEEcompsocitemizethanks{\IEEEcompsocthanksitem Haoning Wu is
with S-Lab, Nanyang Technological University, Singapore. E-mail: haoning001@e.ntu.edu.sg. \protect}
\IEEEcompsocitemizethanks{\IEEEcompsocthanksitem Weisi Lin is with the School of Computer Science and Engineering, Nanyang Technological University, Singapore.  E-mail: wslin@ntu.edu.sg.\protect}}
% \IEEEcompsocitemizethanks{\IEEEcompsocthanksitem 
% $^{\dagger}$Corresponding Authors. \protect}}
        % <-this % stops a space
% \thanks{This paper was produced by the IEEE Publication Technology Group. They are in Piscataway, NJ.}% <-this % stops a space
% \thanks{Manuscript received April 19, 2021; revised August 16, 2021.}}

% The paper headers
% \markboth{Journal of \LaTeX\ Class Files,~Vol.~14, No.~8, August~2021}%
% {Shell \MakeLowercase{\textit{et al.}}: A Sample Article Using IEEEtran.cls for IEEE Journals}

% \IEEEpubid{0000--0000/00\$00.00~\copyright~2021 IEEE}
% Remember, if you use this you must call \IEEEpubidadjcol in the second
% column for its text to clear the IEEEpubid mark.

\maketitle

\begin{abstract}
Image quality scoring and interpreting are two fundamental components of Image Quality Assessment (IQA). The former quantifies image quality, while the latter enables descriptive question answering about image quality. Traditionally, these two tasks have been addressed independently. However, image-quality-specific psychophysical studies suggest that these two tasks are conceptually interconnected: interpreting explicitly represents perceived quality attributes whereas scoring summarizes such evidence into an overall quality judgment. Thus, unifying these capabilities within a single model is both intuitive and logically coherent.
In this paper, we propose \textbf{\textsc{Q-SiT}} (\underline{Q}uality \underline{S}coring and \underline{I}nterpreting joint \underline{T}eaching), a unified framework that enables large multimodal models (LMMs) to learn both image quality scoring and interpreting simultaneously. We achieve this by transforming conventional IQA datasets into learnable question-answering datasets and incorporating human-annotated quality interpreting data for training. Furthermore, we introduce an efficient scoring \& interpreting balance strategy, which first determines the optimal data mix ratio on lightweight LMMs and then maps this ratio to primary LMMs for fine-tuning adjustment. This strategy not only mitigates task interference and enhances cross-task knowledge transfer but also significantly reduces computational costs compared to direct optimization on full-scale LMMs.
With this joint learning framework and corresponding training strategy, we develop \textbf{\textsc{Q-SiT}}, the first model capable of simultaneously performing image quality scoring and interpreting tasks, along with its lightweight variant, \textbf{\textsc{Q-SiT-mini}}. Experimental results demonstrate that \textbf{\textsc{Q-SiT}} achieves strong performance in both tasks with superior generalization IQA abilities, while \textbf{\textsc{Q-SiT-mini}} significantly reduces computational overhead while maintaining competitive performance. Project page at \url{https://github.com/Q-Future/Q-SiT}.
\end{abstract}


\section{Introduction}
\begin{figure}
    \centering
    \includegraphics[width=0.8\linewidth]{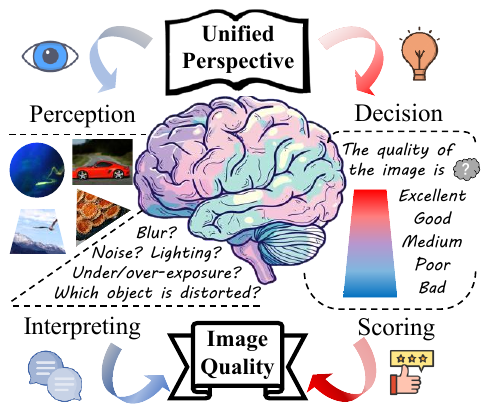}
    \caption{The perception--decision perspective \cite{leisti2009subjective,leisti2022fewer} for image quality scoring and interpreting tasks recognizes that these two processes are deeply interconnected. While most previous approaches treat them separately with distinct models, \textbf{interpreting} (\textit{the perceptual process}) and \textbf{scoring} (\textit{the decision-making process}) are not independent from the human vision system (HVS). Instead, they are integral components of a unified evaluation framework.}
    \label{fig:spotlight}
    \vspace{-0.5cm}
\end{figure}

\IEEEPARstart{T}{raditional} Image Quality Assessment (IQA), as a leading task in low-level vision, primarily focuses on quantifying image quality scores to guide the optimization of image reconstruction, enhancement, and generation tasks, which serves as a crucial metric for compression and transmission systems as well.
With the advancement of Large Multimodal Models (LMMs), IQA has become more interpretable, making the evaluation process more transparent. Unlike conventional approaches that directly output numerical scores, LMMs can answer quality-related questions or describe image quality in natural language.
Given that image quality scoring involves producing numerical ratings, while image quality interpreting entails generating textual responses, most existing work treats these two tasks separately with different models \cite{q-instruct,qalign,depictqa_v1,depictqa_v2}. However, from the perspective of human vision, these two tasks are inherently interconnected. Image-quality-specific psychophysical studies suggest that subjective quality evaluation involves both describing perceived quality attributes~\cite{leisti2009subjective,virtanen2020underlying} and forming an overall quality decision from such subjective experience~\cite{leisti2022fewer}. Therefore, \textbf{interpreting} (\textit{the perceptual process}) and \textbf{scoring} (\textit{the decision-making process}) are not separate but rather part of a unified evaluation process as shown in Fig. \ref{fig:spotlight}. Therefore, a natural insight follows:

\textit{Teaching LMMs to perform both image quality scoring and interpreting tasks within a unified framework is not only feasible but also a necessary step forward.}

\begin{figure*}
    \centering
    \includegraphics[width=.95\linewidth]{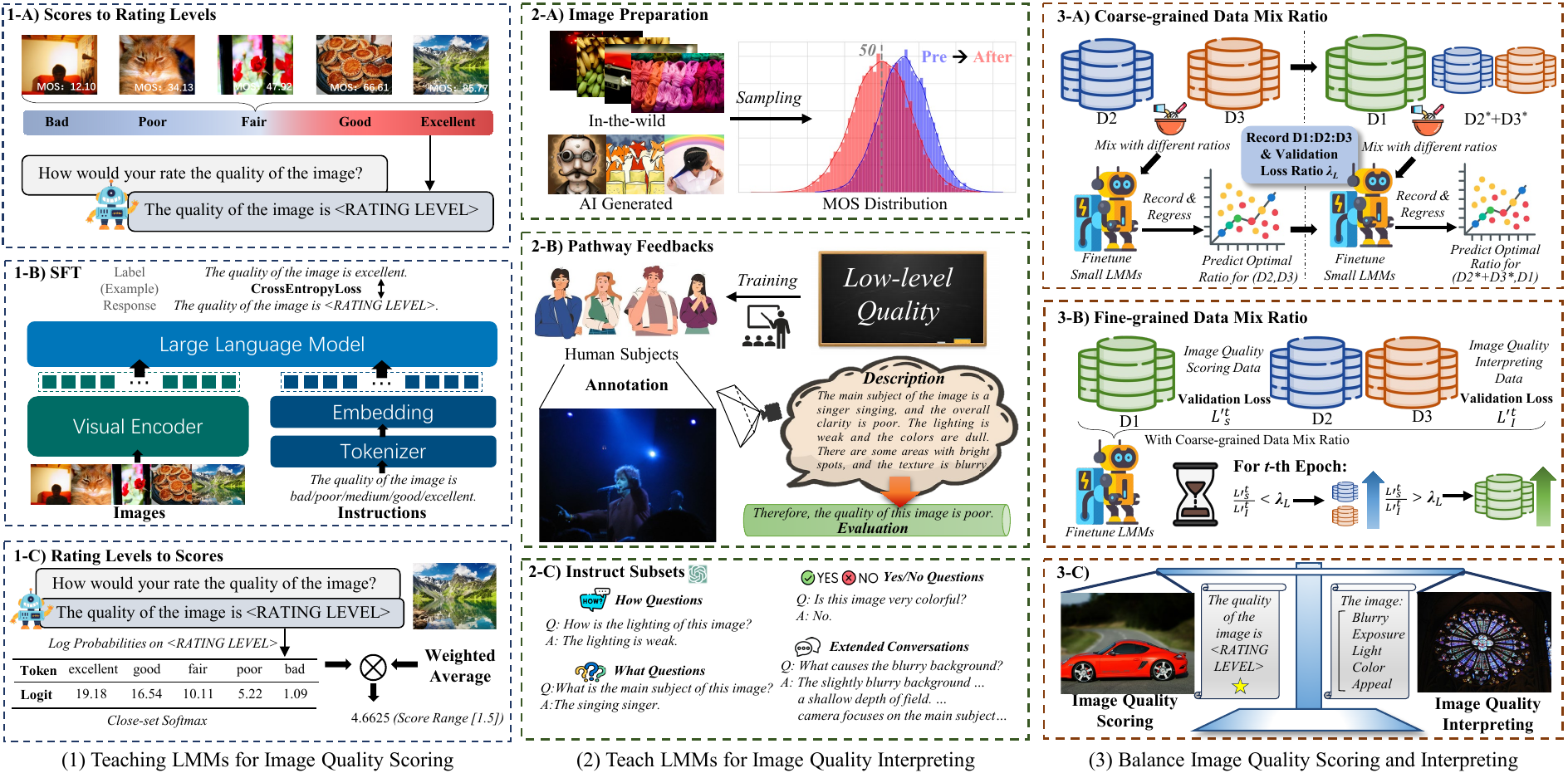}
    \caption{Overview of the proposed \textbf{\textsc{Q-SiT}} framework. The image quality scoring framework (\textbf{Part1}) trains LMMs to evaluate images by assigning probabilities to text-defined quality levels. The image quality interpreting framework (\textbf{Part2}) incorporates large-scale human annotations to inject low-level knowledge into the LMMs. Additionally, we introduce a balance strategy (\textbf{Part3}) to regulate the data ratio, ensuring robust performance across both tasks.}
    \label{fig:framework}
    \vspace{-0.2cm}
\end{figure*}

Therefore, to fulfill this insight, we focus on 3 major parts in this work: \textbf{1}) teach LMMs for image quality scoring, \textbf{2}) teach LMMs for image quality interpreting, and \textbf{3}) balance scoring and interpreting, enabling the model to bridge the gap and efficiently leverage multi-task knowledge.
\newline \newline
\noindent \textbf{Part1: \textit{How to teach LMMs for image quality scoring?}} 
To design the most effective syllabus, we first review the standard process for collecting MOS from human annotators~\cite{itu}. Typically, organizers define several rating levels (\textit{e.g.}, \textit{excellent}, \textit{fair}, \textit{bad}) and select representative examples for each level to align annotators with the rating standards. Based on these predefined levels, human annotators provide their ratings either by selecting a categorical option or using a grade-guided slider. In other words, \textbf{human annotators never learn or assign a precise numerical score} (\textit{e.g.}, \textit{3.457} on a [1,5] scale). Instead, these final scores are computed from the distribution of human ratings.
Meanwhile, recent studies~\cite{wu2024qbench,qalign} have observed that LMMs exhibit \textit{behavioral patterns} similar to those of human annotators when instructed to score images. Specifically, they tend to {respond with text-defined levels} (\textit{e.g.}, \textit{good} or \textit{poor}). Even when explicitly prompted to provide numerical scores, their accuracy is significantly lower compared to deriving scores from categorical levels. Therefore, directly training LMMs to output raw scores may not be optimal.

Given these observations, we propose a \textit{human-emulating syllabus} for training LMMs in visual scoring as illustrated in Fig.\ref{fig:framework}-(1). To mimic the process of training human annotators during training, we convert MOS values into five text-defined rating levels\cite{itu} (\textit{excellent, good, fair, poor, bad}), which are then formatted into instruction-response pairs to facilitate visual instruction tuning~\cite{llava} on LMMs. {During inference}, replicating the MOS collection strategy used for human ratings, we extract the log probabilities of different rating levels, apply softmax pooling to obtain the closed-set probabilities for each level, and compute the final LMM-predicted score via a weighted average of these probabilities.
\newline \newline
\noindent \textbf{Part2: \textit{How to teach LMMs for image quality interpreting?}}  
Image quality is closely related to low-level features. However, a key challenge is the lack of low-level visual datasets for training LMMs, as publicly available datasets typically focus on high-level visual abilities~\cite{okvqa,cocovqa,gqa,vg}. To address this issue, we construct a large-scale \textit{low-level visual instruction tuning SFT} dataset, designed to inject low-level knowledge into LMMs. This process occurs in two steps:

\textit{Step 1: Collect human feedback on low-level vision.}

In this step, we invite human participants to provide direct feedback on their low-level perception and understanding of a variety of images. Specifically, each feedback consists of two parts: \textbf{1}) An exhaustive {description} of elemental low-level attributes (\textit{e.g., blurs, noise, clarity, color, brightness}). These descriptions also incorporate context related to content\cite{sfa,rfugc} or position~\cite{wu2022fastervqa,paq2piq} (\textit{e.g., {the duck / the left part of the image} is underexposed}). \textbf{2}) An overall conclusion on the image quality based on the attribute descriptions. Together, these two components can be represented as pathway feedbacks for both fundamental human perception and the reasoning process behind visual quality evaluation. The resulting dataset contains 58K pathway feedbacks for 18,973 multi-sourced images, each with at least three feedbacks (\textit{avg. 46.4 words per feedback}).

\textit{Step 2: Convert these feedbacks for instruction tuning.}

While the \textit{pathway} feedbacks form a crucial subset for \textit{low-level visual instruction tuning}, the full instruction tuning dataset should be designed to activate broader capabilities. Specifically, it should also include a low-level \textit{visual question answering} (VQA) subset. To create a reliable VQA subset, we follow approaches used for COCO-VQA~\cite{cocovqa}, where image captions are converted into question-answer pairs. We employ GPT~\cite{chatgpt} to transform the \textit{pathway} feedbacks into question-answer pairs, with adjectives (\textit{e.g., good/fair/poor}) or nouns (\textit{e.g., noise/motion blur}) as answers. Additionally, we create a balanced \textit{yes-or-no} question-answer set, based on the information provided in the feedbacks (\textit{answered with yes}) or contrasting information (\textit{answered with no}). Some context-related question-answer pairs are also developed to better ground~\cite{refcoco} low-level attributes. Following established practices~\cite{AOKVQA}, all question-answer pairs in the VQA subset include both \textit{yes-or-no} questions (\textit{Yes/No}) and multiple-choice \textit{what/how} questions (\textit{A/B/C/D}). Furthermore, with the assistance of GPT, we collect a subset of long conversations related to low-level concerns (\textit{e.g., why distortions occur}, \textit{how to improve image quality}). These subsets combine to form the image quality interpreting dataset as shown in Fig.~\ref{fig:framework}-(2-C), containing 200K instruction-response pairs, designed to enhance LMMs' performance on low-level visual tasks. \newline \newline
\noindent \textbf{Part3: \textit{How to balance image quality scoring and interpreting?}}  
We would like to present an interesting and common phenomenon as shown in Fig.~\ref{fig:radar}. When LMMs are trained on the quality scoring task, their ability to perform the interpreting task significantly degrades (\textit{Q-Align}~\cite{qalign}). However, when trained on the quality interpreting task, their performance on the scoring task improves to some extent (\textit{Q-Instruct}~\cite{q-instruct}). This phenomenon arises due to two main reasons:
\textbf{1}) Differences in \textbf{task requirements}: Interpreting tasks require more detailed and comprehensive perceptual abilities, whereas scoring tasks need a certain level of abstract summarization skills. If an LMM bypasses the comprehensive perception and directly learns the summarization skills required for scoring, its ability to interpret visual information will significantly degrade.
\textbf{2}) Differences in \textbf{task formats}: The scoring task ultimately requires a quantifiable quality score, typically achieved through direct quality-related questions. On the other hand, interpreting tasks include a variety of question formats, such as \textit{yes-or-no}, \textit{what/how}, and \textit{open-ended} questions. The format of the scoring task can be seen as a subset of the interpreting task's format. Therefore, focusing solely on interpreting tasks can, to some extent, improve performance on the scoring task.

Although image quality scoring and interpreting are closely related tasks, their differences in required abilities and task formats (\textit{quantification vs. natural language description}) make it impossible to maintain the performance of a single task when simply combining and mixing data for joint training. Furthermore, training exclusively on low-level, strongly correlated data will degrade the LMM's semantic understanding, which in turn can reduce its interpreting abilities (since interpreting also involves detecting and recognizing contextual information).
To address these issues in multi-task training, we have designed two strategies:
\textbf{1}) Incorporating common VQA data to maintain the LMM's understanding of general semantic information.
\textbf{2}) Introducing a dynamic dataset mixing strategy to adaptively select the optimal proportion of different datasets for training.

More specifically, we classify the data into three types: image quality scoring data (\textbf{D1}), image quality interpreting data (\textbf{D2}), and image general understanding data (\textbf{D3}). First, we train a small LMM on various data mixing ratios and construct regression models based on the resulting performance to identify the optimal data mixing ratio to save computational resources\cite{liu2024regmix}. Using this ratio, we train the small LMM and obtain the lightweight \textbf{\textsc{Q-SiT-mini}} with the validation set loss. We then use this ratio as well as the loss information to sample data for training the primary \textbf{\textsc{Q-SiT}}. In this way, we can not only mitigate task interference and enhance cross-task knowledge transfer but also significantly reduces computational costs compared to direct optimization on full-scale LMMs.

This work is a significantly extended version of our previous publications \cite{qalign} and \cite{q-instruct}. Compared to these earlier works, we introduce three major improvements:
\textbf{1}) Most importantly, we propose \textbf{a unified framework} that enables a single LMM to perform both quality scoring and interpreting, bridging the gap between these tasks, which were previously handled by separate models.
\textbf{2}) For the joint learning framework of quality scoring and interpreting, we introduce \textbf{a training strategy} to control the proportion of mixed datasets. This strategy helps efficiently save computational resources in searching for the optimal settings and effectively mitigates conflicts arising from different tasks.
\textbf{3}) Additionally, we propose \textbf{a lightweight model} that significantly reduces inference costs and time while preserving competitive performance.

In summary, we explore the LMM joint teaching framework for image quality scoring and interpreting. The unified framework not only follows the perception-decision view in psychology but also addresses the practical need for both scoring and interpreting simultaneously. Our contributions can be summarized as follows:
\begin{itemize} 
\item We develop \textbf{pipelines} for teaching LMMs image quality scoring and interpreting. The quality scoring pipeline trains LMMs to score images based on probabilities corresponding to text-defined quality levels, while the quality interpreting pipeline leverages large-scale human annotations to inject low-level knowledge into LMMs.
\item We introduce a \textbf{unified framework} that enables a single LMM to perform both quality scoring and interpreting tasks. Additionally, we propose an efficient \textbf{balance strategy} that optimizes data ratio selection while reducing computational costs. This strategy mitigates potential performance degradation and ensures the LMM maintains strong performance across both tasks.
\item We also propose a \textbf{lightweight} LMM model that significantly reduces inference costs and time while maintaining good performance across scoring and interpreting tasks, making it a promising solution for mobile deployment.
\end{itemize}

\section{Related Works}

\subsection{Large Multi-modality Models}
Large language models (LLMs), such as GPT-4~\cite{openai2023gpt4}, T5~\cite{flant5}, and LLaMA~\cite{llama}, demonstrate remarkable linguistic capabilities across general human knowledge domains. By integrating visual input via CLIP~\cite{CLIP} and incorporating additional adaptation modules, large multimodal models (LMMs)~\cite{otter,llamaadapterv2,llava,iblip,xcomposer} have been developed to tackle a wide range of multimodal tasks, including image captioning, visual question answering, segmentation, classification, and reasoning.
OpenFlamingo~\cite{openflamingo} pioneered this approach by integrating gated cross-attention dense blocks into pretrained language encoder layers. InstructBLIP~\cite{iblip} further extended BLIP-2~\cite{blip2} through vision-language instruction tuning. To advance the development of open-source LMMs, numerous studies have leveraged GPT-4~\cite{openai2023gpt4} to generate vision-language training data, leading to the creation of the LLaVA series~\cite{llava,improvedllava,liu2024llavanext}.

\begin{figure}
    \centering
    \includegraphics[width=0.88\linewidth]{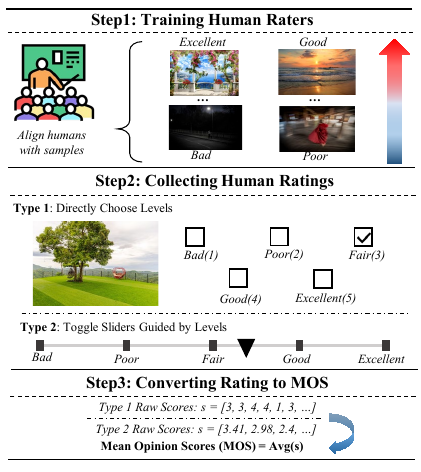}
    \caption{This illustration outlines the human annotation process, which typically involves three stages:
\textbf{1}) Training human raters with text-defined rating levels. To simulate this, we propose a rating-level-based syllabus tailored for LMMs.
{\textbf{2})} Collecting human ratings. Raters either select a level (\textbf{Type 1}) or adjust a level-guided slider to score (\textbf{Type 2}), \textit{without directly entering the score in either method}.
{\textbf{3})} Converting initial ratings to MOS through a weighted average. In this final stage, we propose a probability-based inference approach for LMMs to predict the final scores.}
    \label{fig:human}
    \vspace{-10pt}
\end{figure}

\subsection{Image Quality Assessment (IQA).}
Image Quality Assessment (IQA) primarily focuses on evaluating the impact of distortions and other quality degradations on human perception. Early IQA algorithms relied on handcrafted features and statistical models~\cite{ssim,brisque,niqe}. However, as distortions become more diverse and visual content grows increasingly complex, data-driven deep neural networks have gained prominence in the field. Notable examples include NIMA~\cite{nima}, DBCNN~\cite{dbcnn}, and HyperIQA~\cite{hyperiqa}, which leverage end-to-end learning to improve IQA performance. Following this trend, MUSIQ~\cite{musiq} introduces a multi-scale input structure with transformers, further enhancing IQA accuracy.

% Despite these advancements, most CLIP-based methods rely on visual-text similarity to predict quality scores, making them slightly less effective than purely visual approaches in some scenarios.

More recently, several approaches have explored vision-language representations from CLIP~\cite{CLIP} to improve text-vision alignment generalization in IQA. Among them, CLIP-IQA+\cite{clipiqa} employs a few-shot learning strategy via CoOp\cite{coop}, while LIQE~\cite{liqe} integrates multitask learning to enhance performance.  With the advancement of LMMs, researchers have increasingly leveraged them for IQA, primarily by estimating the probabilities of adjective-based quality ratings and converting these probabilities into numerical scores~\cite{qalign,zhang2024q}.
Furthermore, many studies have begun to focus on leveraging LMMs for image quality interpreting, leading to the creation of various multimodal datasets designed to inject quality-related knowledge into LMMs~\cite{q-instruct,depictqa_v1,depictqa_v2}. However, current approaches still treat image quality scoring and interpreting as separate tasks, relying on independent models for each. To date, no unified framework or corresponding training strategy has been proposed to jointly handle both tasks.

\section{The \textbf{\textsc{Q-SiT}}}
In this section, we first introduce the pipelines for image quality scoring and interpreting separately, and then propose a balance strategy for both image quality scoring and interpreting.

\subsection{Teaching LMMs for Image Quality Scoring}
\label{sec:qalign}
\subsubsection{[\textit{Insight 1}] How Do Humans Rate?}
To design a curriculum for training LMMs to score images, we first examine the process of collecting human ratings (Fig.~\ref{fig:human}). This process generally involves three key stages:

\textbf{Stage 1: Training Human Raters.} In line with standard human rating procedures~\cite{itu}, it's crucial to train raters on rating rules. This includes aligning them with examples for each {rating level}. Importantly, the quality scores of these examples are \textit{not revealed} to the human raters during this stage.

\textbf{Stage 2: Collecting Human Ratings.} Once raters are trained, the next step is to gather initial human ratings. Raters typically provide their assessments in two ways: {\textbf{1})} By selecting a rating level directly or {\textbf{2})} By adjusting a slider to generate a score. In both cases, raters are not required to \textit{explicitly input} numerical scores when providing their opinions.

\textbf{Stage 3: Converting Human Ratings to MOS.} As illustrated in Fig.~\ref{fig:human}, initial ratings are averaged into a Mean Opinion Score (MOS) for visual scoring datasets. Importantly, human raters \textit{do not participate} in this final stage.

Throughout all three stages, raters are neither trained nor instructed to predict a numerical score. This approach reflects how, in everyday situations, people typically respond with \textbf{qualitative adjectives} (\textit{e.g., fine, poor}) rather than \textbf{precise numerical ratings} (\textit{e.g., 2.66, 1.22}). By leveraging humans' innate ability to provide qualitative feedback, we reduce cognitive load and improve the accuracy of subjective evaluations.

% Having analyzed the human rating process, we explore the analogous capabilities of LMMs.

\subsubsection{[\textit{Insight 2}] How Do LMMs Rate?}
Since LMMs are fundamentally designed to understand/generate human-like text, they should in theory exhibit behaviors similar to humans. To test this hypothesis, we prompt five LMMs\footnote{None of these LMMs are specifically trained for visual rating tasks.} with the following instructions and analyze their responses:

\textit{{\tt <img>} How would you rate the quality of the image.}

The results, shown in Tab.\ref{tab:lmminnate}, reveal that, before specific alignment, LMMs predominantly output \textbf{qualitative adjectives}. This suggests that, like humans, LMMs initially prefer qualitative feedback over numerical scores. Therefore, to minimize the learning overhead, we choose rating levels as the targets. 

\subsubsection{Conversion Between Rating Levels and Scores}

Building upon the general methodology for training LMMs with rating levels, we explore two key aspects: converting scores from existing datasets into discrete rating levels \textit{during training}, and obtaining scores from LMMs \textit{during inference}.

\subsubsection{[\textit{Training}] Scores $\to$ Rating Levels}

 During the training phase, we map continuous scores to discrete rating levels. Since adjacent human ratings are typically equidistant, we use equidistant intervals for this conversion. Specifically, we divide the range between the highest score ($\mathrm{M}$) and the lowest score ($\mathrm{m}$) into five equal intervals. Scores falling within each interval are then assigned to the corresponding rating level:
\begin{equation}
    {L(s)} \! = \! l_i \text{  if } \text{m}\! + \!\frac{i-1}{5} \!\times \!\mathrm{(M-m)} \!< \!s \! \leq \! \mathrm{m} \!+\! \frac{i}{5} \!\times\! \mathrm{(M-m)},
\end{equation}
where \{$l_i|_{i=1}^{5}\}=\{\textit{bad, poor, fair, good, excellent}\}$ are the standard text rating levels as defined by ITU~\cite{itu}. The conversion mapping $L$ described above is a \textit{multi-to-one} mapping, which inherently leads to a small loss in precision relative to the ground truth. In Tab.~\ref{tab:conversionprecision}, we present the conversion precision across five datasets. The results demonstrate that the conversion maintains a strong linear correlation with the original scores.

\begin{table}[t]\small
    \vspace{-1.2em}
    \centering
    \renewcommand\arraystretch{1.14}
    \renewcommand\tabcolsep{5.5pt}
    \caption{[\textit{Insight 2}] \textsc{How do LMMs rate?} The responses from LMMs to the question “\textit{How would you rate the quality of the image?}” across 1168 images from the LIVEC dataset \cite{livechallenge} indicate that LMMs tend to favor qualitative adjectives in their ratings.}
    \resizebox{\linewidth}{!}{\begin{tabular}{l|ccc}
    \hline
     Model / Frequency & \textbf{Qualitative Adjectives} & \textbf{Numerical Ratings} \\  \hdashline
     Qwen2.5-VL~\cite{Qwen2.5-VL} & \textbf{97\%} (1132/1168) & 4\% (4361168) \\
     LLaVA-v1.5~\cite{improvedllava} & \textbf{100\%} (1168/1168) &  0\% (0/1168)\\
     mPLUG-Owl-3~\cite{ye2024mplugowl3} & \textbf{100\%} (1168/1168) & 0\% (0/1168)\\
     InternVL2.5~\cite{internvl2.5} & \textbf{99\%} (1157/1168) & 1\% (11/1168) \\
     Shikra~\cite{shikra} & \textbf{100\%} (1168/1168) & 0\% (0/1168) \\
     \hline
    \end{tabular}}
    \vspace{-12pt}
    \label{tab:lmminnate}
\end{table}

\subsubsection{[\textit{Inference}] Rating Levels $\to$ Scores}

After the model has been trained, we need to reverse the process and convert the rating levels back into scores. To simulate the post-processing of human ratings (Fig.~\ref{fig:human} Step 3), we first define the reverse mapping $G$ that maps text-defined rating levels back to scores: \begin{equation} G: l_i \rightarrow i \end{equation} For example, the rating \textit{fair} is mapped back to score 3, and \textit{bad} to score 1.
Then the MOSs are calculated as a weighted average of the converted scores, weighted by the frequency $f_{l_i}$:
\begin{equation}
    G: l_i \rightarrow I,
\end{equation}
For instance, \textit{fair} is converted back to score 3, and \textit{bad} to 1. 
In human opinion collection (Fig.~\ref{fig:human} Step 2 \textbf{Type 1}), the MOS values are calculated via the weighted average of the converted scores and frequencies $f_{l_i}$ for each level: $\mathrm{MOS}=\sum_{i=1}^5 f_{l_i} G(l_i)$. Similarly, for LMMs, we substitute the $f_{l_i}$ with the LMM-predicted probabilities for each level. Given that the predicted {\tt<LEVEL>} token of LMMs is the probability distribution (denoted as $\mathcal{X}$) on all possible tokens, we conduct a close-set softmax on \{$l_i|_{i=1}^{5}$\} to get the probabilities $p_{l_i}$ for each level ($p_{l_i}$ for all $l_i$ sum as 1):

\begin{equation}
    p_{l_i} = \frac{e^{\mathcal{X}_{l_i}}}{\sum_{j=1}^{5} {e^{\mathcal{X}_{l_j}}}},
\end{equation}
and the final predicted scores of LMMs are denoted as 
\begin{equation}
    \mathrm{S_{LMM}}=\sum_{i=1}^5 p_{l_i} G(l_i) = i \times  \frac{e^{\mathcal{X}_{l_i}}}{\sum_{j=1}^{5}{e^{\mathcal{X}_{l_j}}}},
    \label{eq:5}
\end{equation}

The inference conversion is theoretically equivalent to the MOS collection process from a set of human ratings in levels. Moreover, it represents the general expression form of the binary softmax strategy ($\mathrm{S_\text{Q-Bench}}=\frac{e^{\mathcal{X}_\textit{good}}}{{e^{\mathcal{X}_\textit{good}}}+{e^{\mathcal{X}_\textit{poor}}}}$) as proposed by \cite{wu2024qbench}, which can be considered as a simplified version of Eq.~\ref{eq:5} with only two rating levels.

\begin{table}[tbp]
\vspace{-1.3em}
\centering
\small
\renewcommand\arraystretch{1.3}
\renewcommand\tabcolsep{3.5pt}
\caption{Precision of training conversion (\textit{Score $\to$ Rating Levels}) on five IQA datasets with MOS. Metrics are SRCC/PLCC.}
\resizebox{\linewidth}{!}{\begin{tabular}{l:ccccc}
\hline
Conversion & \textbf{KonIQ} & \textbf{SPAQ} & \textbf{LIVEC} & \textbf{AGIQA-3K} & \textbf{KADID}  \\ \hdashline
\textit{Scores $\to$ Levels} & 0.952/0.961 & 0.969/0.968  & 0.933/0.930 & 0.942/0.955 & 0.979/0.982\\
\hline
\end{tabular}}
\label{tab:conversionprecision}
\vspace{-12pt}
\end{table}

% After preparing the images, we explain the rationale behind and provide a detailed task definition for the \textit{pathway} feedback, which consists of natural language responses.

\subsection{Teach LMMs for Image Quality Interpreting}
\label{sec:qins}
Training LMMs for image quality interpreting heavily relies on human annotations. \textbf{1}) First, we introduce the \textit{pathway} feedback dataset, a large-scale collection of detailed {text} feedbacks from humans on low-level visual aspects. To ensure diversity and balance in different low-level appearances, we subsample images from diverse sources and reduce biases in the source distributions.  \textbf{2}) We then design additional instruction types to enable LMMs to handle a variety of human queries. These include a \textit{visual question answering} subset to improve low-level perception accuracy, and an \textit{extended conversation} subset that allows LMMs to seamlessly \textit{chat} with humans about topics related to low-level visual aspects.

\subsubsection{Preparation of Images}
The images are sampled from a variety of sources, including four \textit{in-the-wild} IQA datasets~\cite{spaq,hosu2020koniq,paq2piq,clive}, and two datasets featuring \textit{AI-generated} images~\cite{agiqa3k,imagereward}. As shown in Tab.~\ref{tab:sub_sample}, the sub-sampled image population is carefully curated to introduce a broader range of low-level appearances, ensuring a balance between high-quality and low-quality images.
To further diversify the low-level appearances, we design a custom variant of \textit{imagecorruptions}~\cite{imagecorruptions} to randomly introduce one of fifteen artificial distortions to 1,012 originally pristine images from the COCO~\cite{cococaps} dataset. The resulting sub-sampled dataset consists of 18,973 images, which are then presented to human subjects for \textit{pathway} feedback.

\subsubsection{Task Definition of Pathway Feedbacks}

For the pathway feedbacks, we aim to collect a richer and more nuanced understanding of human perception of low-level visual aspects. Rather than gathering multi-dimensional scores, as in existing studies~\cite{spaq,wu2023explainable}, we introduce a new format of annotation that includes a comprehensive natural language description of low-level visual attributes (\textit{e.g., noise, brightness, clarity}), followed by a general conclusion. The rationale behind this format is as follows:
First, these descriptions better preserve human perceptions in a more \textit{complete} and \textit{precise} manner. For instance, if an image contains both dark and bright areas, a simple brightness score may fail to capture the full context~\cite{paq2piq,wu2022fastervqa}. The positional context may be lost, and the score’s reliability could be compromised, as neither labeling the image as `\textit{dark nor bright}' would be entirely accurate.
Additionally, unlike free-form text feedback, the structure of pathway feedback generally mirrors the human reasoning process. For example, when shown an underexposed yet clear image, human subjects often offer intuitive reasoning that leads to conclusions such as: `\textit{Thus, the quality of the image is acceptable}.' This reasoning process helps LMMs better align with human perception and understanding of low-level visual attributes. While transforming this \textit{pathway}-style format into machine learning objectives posed challenges in the past, the advent of LMMs provides an opportunity to learn directly from these human \textit{pathway} feedbacks, enabling machines to more precisely and robustly emulate human perception.

\subsubsection{The subjective study process.} 

The subjective study is conducted in a well-controlled laboratory setting, with a total of 39 \textit{trained} human subjects invited to participate. The training material not only includes calibration for \textit{overall quality}, but also provides guidance on the \textit{respective text descriptions} of various low-level visual attributes. Since most of the images are sourced from IQA datasets, the mean opinion scores (MOSs) for these images are also shown to the subjects to help calibrate their understanding of \textit{quality}. To support their feedback process, we present a reference attribute set that can be used in the descriptions. To prevent test fatigue, subjects are warned if they provide feedback on more than 30 consecutive images, and participation is paused after 50 images. In total, 58K pathway feedbacks are collected during the study.

\begin{table}[!t]\small
    \centering
\renewcommand\arraystretch{1.10}
\renewcommand\tabcolsep{4pt}
    \caption{We subsample the source images to mitigate biases in their $\mathrm{MOS}$ distributions, resulting in a more balanced sampled distribution. }
    \vspace{-6pt}
   \resizebox{\linewidth}{!}{\begin{tabular}{l|ccc|ccc}
    \toprule
    \multirow{2}{80pt}{\textbf{Image Sources \textcolor{gray}{$\mathrm{MOS}\in[0,100)$}}} & \multicolumn{3}{c|}{Original Distribution} & \multicolumn{3}{c}{Sampled Distribution}  \\ \cline{2-7}
      & {Size} & {$\mu_{\mathrm{MOS}}$} & $\sigma_{\mathrm{MOS}}$ & {Size} & {$\mu_{\mathrm{MOS}}$} & $\sigma_{\mathrm{MOS}}$\\ \hline
     KonIQ~\cite{hosu2020koniq} & 10,073 & 58.73  & 15.43  & 5,182 & 49.53 & 15.72\\
     SPAQ~\cite{spaq} & 11,125 & 50.32 & 20.90 & 10,797 & 49.46 & 20.63\\
     LIVE-FB~\cite{paq2piq} & 39,810 & 72.13 & 6.16 & 800 & 60.68 & 17.38  \\
     LIVE-itw~\cite{clive} & 1,169 & 55.38 & 20.27 & 200 & 55.70 & 19.83  \\ 
     AGIQA-3K~\cite{agiqa3k} & 2,982 & 50.00 & 19.80 & 400 & 40.80 & 21.80 \\
     ImageRewardDB~\cite{imagereward} & 50,000 & \multicolumn{2}{c|}{\textit{- w/o $\mathrm{MOS}$ -}} & 584 & \multicolumn{2}{c}{\textit{- w/o $\mathrm{MOS}$ -}}\\
     15\textit{-distortion} COCO~\cite{cococaps} & 330,000 &  \multicolumn{2}{c|}{\textit{- w/o $\mathrm{MOS}$ -}} & 1,012 & \multicolumn{2}{c}{\textit{- w/o $\mathrm{MOS}$ -}}   \\ \hline
     \textit{Overall}  & 445,159 & 65.02 & 16.51 & 18,973 & \underline{49.87} & \underline{19.08}  \\
    \bottomrule
\end{tabular}
 }
\vspace{-12pt}
    \label{tab:sub_sample}
\end{table}

\subsubsection{Visual Question Answering (VQA)}

In addition to directly applying the pathway feedbacks for low-level visual instruction tuning, we also design a pipeline involving GPT~\cite{chatgpt} to convert these feedbacks into a \textit{visual question answering} (VQA) subset. Specifically, we ask GPT to generate diverse questions related to low-level vision from the \textit{pathway} feedbacks and provide answers using \textit{as few words as possible}. Through this process, we transform the feedbacks into {76K} questions, including \textit{how} questions answered with opinion-related adjectives (\textit{e.g., good/poor, high/low}), and \textit{i.e.} \textit{what} questions answered with attribute-related nouns (\textit{e.g., blur/noise/focus}) or context-related nouns (\textit{e.g., left/the peacock/the background}).
We also instruct GPT to generate binary judgments (\textit{yes/no}) from the feedbacks, ensuring a balanced 1:1 ratio of \textit{yes} and \textit{no}, resulting in {57K} \textit{yes-or-no} questions. For the answering format, following the A-OKVQA approach~\cite{AOKVQA}, we not only provide the correct answers but also create several distracting options, transforming them into an additional multiple-choice question (MCQ) format.

\subsubsection{Extended Conversations}

While the first two subsets are designed to enhance fundamental language-related abilities for low-level vision, the third subset, \textit{extended conversations}, focuses on improving the ability to engage in discussions on the low-level visual aspects of an input image. These discussions cover five key areas: a) Analyzing the causes of low-level visual patterns; b) Offering suggestions for improving photography; c) Recommending tools for restoring, enhancing, or editing the image; d) Suggesting the image to relevant consumers; e) Engaging in other relevant conversations based on the low-level visual descriptions provided in the pathway feedbacks. Like the other subsets, the extended conversation subset is also generated by GPT, with a total of {12K} conversations collected.

\begin{algorithm}[t]\small
\caption{Optimal Data Mix Ratio Estimation}
\label{alg:data_mix}
\begin{algorithmic}[1]  % [1] 使行号正常从 1 开始
\Require Initial datasets [\textbf{D1},\textbf{D2},\textbf{D3}], Validation Set \textbf{V}
\Ensure Optimal mix ratio 
\State \textbf{Step 1: Optimization for D2 \& D3}
\State Initialize search range for $r^{D2:D3}$
\For {$r_A \in \text{Range} $}
    \State Train small model on $(D2_{r_A}, D3_{1-r_A})$
    \State Evaluate performance on validation set \textbf{V}
    \State Store $(r, \text{Performance})$
\EndFor
\State Fit regression model $\mathbb{A}(\cdot)$ to determine:

$r^{D2:D3} = \arg\max_{r_A} \mathbb{A}(D2_r, D3_{1-r})$

\State \textbf{Step 2: Incorporate D1}
\State Initialize search range for $r^{|D2^*+D3^*|:D1}$
\For {$r_B \in \text{Range}$}
    \State Train small model on $(|D2^*\!\!+\!\!D3^*|_{r_B}, D1_{1-r_B})$
    \State Evaluate performance on validation set $V$
    \State Store $(r_B, \text{Performance})$
\EndFor
\State Fit regression model $\mathbb{B}(\cdot)$ to determine:

$r^{|D2^*\!+\!D3^*|:D1}\!\! = \!\! \arg\max_{r_B} \mathbb{B}(\!(D2^*\!\!+\!D3^*)_{r_B}, D1_{1\!-\!r_B}\!)$
\State Record the coarse-grained ratio of \textbf{D1}:\textbf{D2}:\textbf{D3}
\State Record reference loss ratio:
$\lambda_{\mathcal{L}} = \frac{\mathcal{L}_{\mathcal{S}}}{\mathcal{L}_{\mathcal{I}}}$

\State \textbf{Step 3: Fine-Grained Adjustment}
\State Train primary model with record coarse-grained ratio
\For{$t$-th training epoch}
    \State Compute validation loss ratio: $\frac{\mathcal{L'}^{t}_{\mathcal{S}}}{\mathcal{L'}^{t}_{\mathcal{I}}}$
    \If{$\frac{\mathcal{L'}^{t}_{\mathcal{S}}}{\mathcal{L'}^{t}_{\mathcal{I}}} < \lambda_{\mathcal{L}}$}
        \State Increase $|D2^*\!\!+\!\!D3^*|$ for next epoch
    \ElsIf{$\frac{\mathcal{L'}^{t}_{\mathcal{S}}}{\mathcal{L'}^{t}_{\mathcal{I}}} > \lambda_{\mathcal{L}}$}
        \State Increase $|D1|$ for next epoch
    \EndIf
\EndFor
\end{algorithmic}
\label{al:mix}
\end{algorithm}

\subsection{Balance Image Quality Scoring and Interpreting}

\subsubsection{Why Mixing All Doesn't Work} 
Mixing all data for training seems like a natural approach, but there are certain limitations within the joint learning framework for these two tasks. \textbf{1}) One issue arises from the differing demands of abstract versus concrete abilities. Image scoring is a quantitative decision-making process that doesn’t require an in-depth description or understanding of image quality. On the other hand, image interpreting emphasizes perceiving image quality and capturing fine details. \textbf{2}) Another issue lies in the format of the data. Image scoring typically involves a structure like `-How would you rate the quality of the image? -The quality of the image is good.' whereas image interpreting data includes \textit{yes-or-no}, \textit{what/how}, and \textit{open-ended} questions. 

As a result, the scoring format can be considered a specialized subset of the interpreting format. Therefore, training on interpreting data enables LMMs to develop a certain level of scoring capability. As shown in Fig.\ref{fig:radar}, LMM (\textit{Q-Instruct}), trained exclusively on image quality interpreting tasks, achieves balanced performance across IQA datasets when evaluated on scoring tasks. However, an excessive proportion of image quality scoring data in training can negatively impact the model's ability to handle interpreting tasks. As depicted in Fig.\ref{fig:radar}, LMM (\textit{Q-Align}), which is trained solely on image quality scoring tasks, exhibits a significant decline in performance when applied to image quality interpreting tasks.

\begin{figure*}[t]\small
    \centering
        \includegraphics[width=0.42\linewidth]{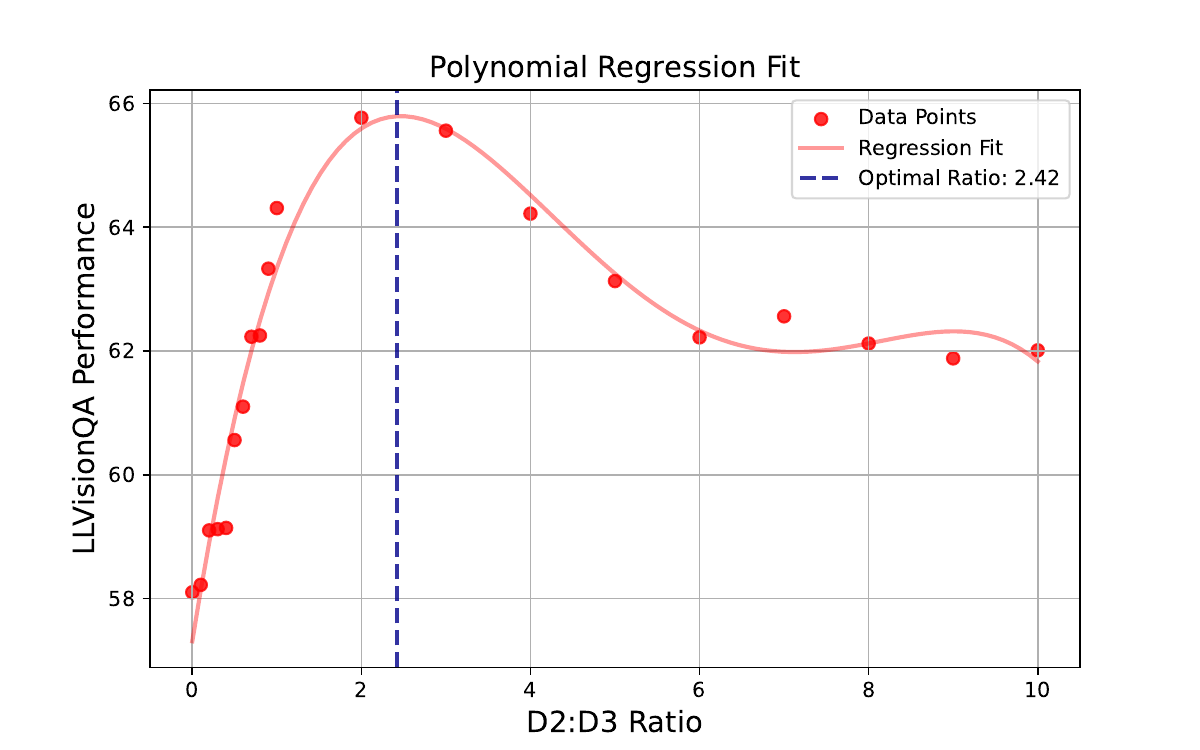}
        \label{fig:d2_d3_ratio}
        \includegraphics[width=0.42\linewidth]{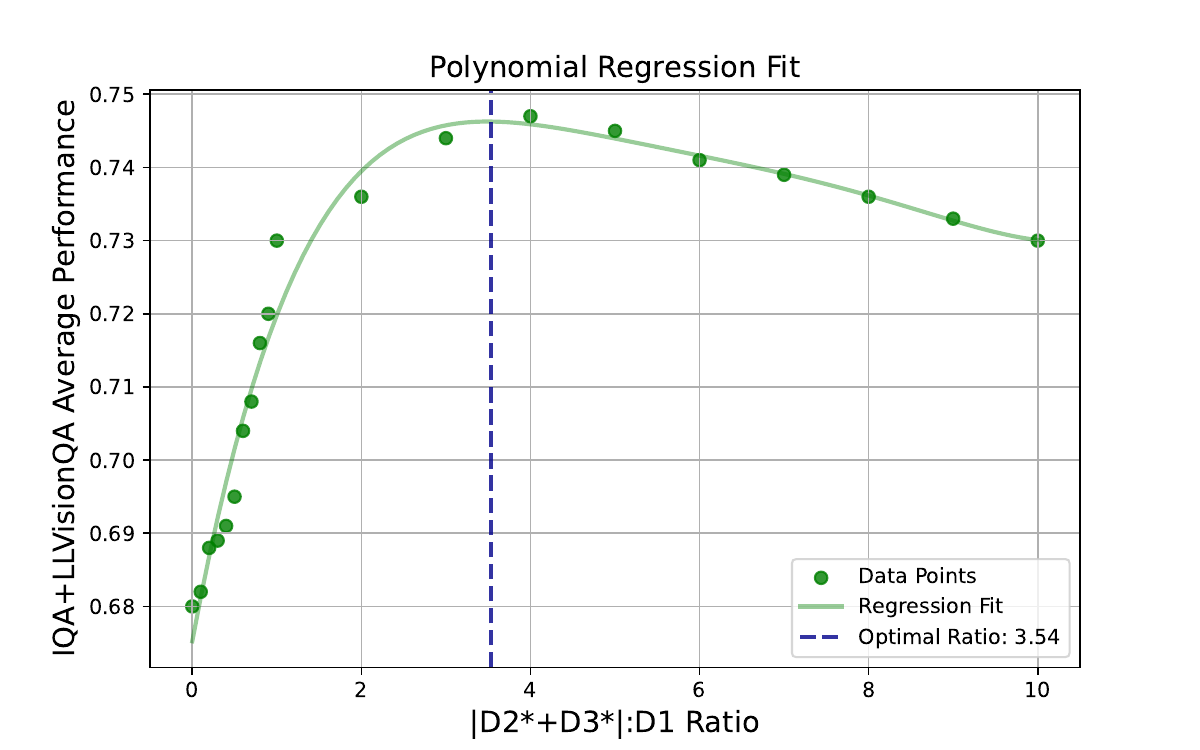}
        \label{fig:d1_d2d3_ratio}
    \caption{Curves for predicting the coarse-grained optimal data mix ratio using a fourth-degree polynomial regression model. Figure \textbf{`D2:D3 Ratio'} illustrates the \textit{LLVisionQA} accuracy trend as the ratio changes. Figure \textbf{`$|$D2$^*$+D3$^*$$|$:D1 Ratio'} depicts the average performance trend across both tasks (mean of \textit{LLVisionQA} accuracy and IQA correlation metrics $\frac{SRCC+PLCC}{2}$), where \textbf{`$|$D2$^*$+D3$^*$$|$'} indicates the \textbf{D2\&D3} mixture with the previously predicted optimal ratio.}
    \label{fig:data_mix_ratios}
    \vspace{-12pt}
\end{figure*}

\subsubsection{Data Utilization}
We use three types of datasets:
\begin{itemize}
    \item We employ modified versions of traditional quality assessment datasets, reformatted into a question-answer structure, to enhance LMMs' image scoring capabilities. This dataset is denoted as \textbf{D1} (\textit{described in Section~\ref{sec:qalign}}). 
    \item To further improve LMMs' image interpreting skills, we utilize the constructed image quality interpreting dataset, referred to as \textbf{D2} (\textit{described in Section~\ref{sec:qins}}).
    \item A significant challenge arises when training solely on low-level data: the semantic understanding of LMMs deteriorates substantially \cite{co_instruct}, directly impacting their ability to interpret images. Fine-grained quality descriptions often require local in-context comprehension. Therefore, we incorporate the LLaVA-Instruct-150K dataset \cite{llava} to preserve semantic cognition of LMMs, denoted as \textbf{D3}.
\end{itemize}
It is important to note that \textbf{D1} follows a relatively fixed question-answer format, which may interfere with the image quality interpreting task. To address this, we prepend a system prefix (\textit{Assume you are an image quality evaluator}) during training with \textbf{D1}, enabling LMMs to better distinguish between the scoring and interpreting tasks.

\begin{table*}[tbp]
\vspace{-1em}
\centering
\small
\renewcommand\arraystretch{1.15}
\renewcommand\tabcolsep{8pt}
\caption{\textbf{\textsc{Q-SiT}}'s performance on image quality scoring task. We adopt KonIQ and SPAQ (\textit{both in-the-wild photography}) as training sets and evaluate on a wide range of test sets. The `Avg.' performance is calculated as the average values of (SRCC+PLCC)/2 across all IQA datasets. The cross-set evaluations are labeled with $^\textsc{Cross}$. Best in \textbf{bold} and second \underline{underlined}}
\resizebox{\linewidth}{!}{\begin{tabular}{l|ccccccccccc}
\hline
\hline
\makebox[0.3\textwidth][l]{\textit{Training Set:}  \textbf{KonIQ}\textit{$\to$Testing Set:}} & \multicolumn{2}{c}{\textbf{KonIQ}} & \multicolumn{2}{c}{\textbf{SPAQ}$^\textsc{Cross}$} &\multicolumn{2}{c}{\textbf{LIVEC}$^\textsc{Cross}$} &\multicolumn{2}{c}{\textbf{AGIQA-3K}$^\textsc{Cross}$} &\multicolumn{2}{c}{\textbf{KADID-10k}$^\textsc{Cross}$}  &\multirow{2}{*}{\textbf{Avg.}} \\ \cline{1-11}
\textbf{Method} &SRCC&PLCC & SRCC&PLCC & SRCC&PLCC & SRCC&PLCC & SRCC&PLCC &  \\
\hline
NIMA (TIP 2018) \cite{nima} & 0.859 & 0.896 & 0.856 & 0.838 & 0.771 & 0.814 & 0.654 & 0.715 & 0.535 & 0.532 & 0.747\\
DBCNN (TCSVT 2020) \cite{dbcnn} & 0.875 & 0.884 & 0.806 & 0.812 & 0.755 & 0.773 & 0.641 & 0.730 &  0.484 & 0.497 & 0.726\\
HyperIQA (CVPR 2020) \cite{hyperiqa} & 0.906 & 0.917 & 0.788 & 0.791 & 0.749 & 0.772 & 0.640 & 0.702 &0.468 & 0.506 & 0.724\\
MUSIQ (ICCV 2021) \cite{musiq} & \underline{0.929} & {0.924} & 0.863 & {0.868} & 0.830 & 0.789 & 0.630 & 0.722 & 0.556 & 0.575 & 0.769\\
CLIP-IQA+ (AAAI 2023) \cite{clipiqa} & 0.895 & 0.909 &  0.864 & 0.866 & 0.805 & 0.832 & 0.685 & 0.736 &  0.654 & 0.653 & 0.790\\
LIQE (CVPR 2023) \cite{liqe} & {0.928} & 0.912 & 0.833 & 0.846 & \underline{0.870} & 0.830 & 0.708 & {0.772} &  {0.662} & {0.667}  & 0.803 \\
Q-Align (ICML 2024) \cite{qalign} & \textbf{0.940} & \textbf{0.941} & \underline{0.887} & \underline{0.886} & {0.860} & {0.853} & {0.735} & {0.772} & {0.684} & {0.674} & \underline{0.823}\\ \hdashline
\textbf{\textsc{\textbf{\textsc{Q-SiT-mini}}}} (\textit{Ours})  & 0.913 & 0.877 & 0.883 & 0.868 & 0.864 & \underline{0.859} & \underline{0.775} & \underline{0.824} & \underline{0.751} & \underline{0.681} & 0.820 \\
\textbf{\textsc{Q-SiT}} (\textit{Ours}) & 0.922 & \underline{0.925} & \textbf{0.891} & \textbf{0.896} & \textbf{0.895} & \textbf{0.891} & \textbf{0.791} & \textbf{0.850} & \textbf{0.760} & \textbf{0.693} & \textbf{0.851} \\
\hline
\end{tabular}}
\resizebox{\linewidth}{!}{\begin{tabular}{l|ccccccccccc}
\hline
\makebox[0.3\textwidth][l]{\textit{Training Set:}  \textbf{SPAQ}\textit{$\to$Testing Set:}} & \multicolumn{2}{c}{\textbf{KonIQ$^\textsc{Cross}$}} & \multicolumn{2}{c}{\textbf{SPAQ}} &\multicolumn{2}{c}{\textbf{LIVEC}$^\textsc{Cross}$} &\multicolumn{2}{c}{\textbf{AGIQA-3K}$^\textsc{Cross}$} &\multicolumn{2}{c}{\textbf{KADID-10k}$^\textsc{Cross}$}  &\multirow{2}{*}{\textbf{Avg.}} \\ \cline{1-11}
\textbf{Method}  &SRCC&PLCC & SRCC&PLCC & SRCC&PLCC & SRCC&PLCC & SRCC&PLCC \\
\hline
NIMA (TIP 2018) & 0.733 & 0.788 &  0.907 & 0.910 & 0.733 & 0.785 & 0.534 & 0.630 & 0.399 & 0.480 & 0.690\\
DBCNN (TCSVT 2020)& 0.731 & 0.758 & 0.908 & 0.913 & 0.702 & 0.748 & 0.459 & 0.518 & 0.490 & 0.508 & 0.673 \\
Fang \textit{et al.} (CVPR 2020) & 0.714 & 0.742  &  0.908 &0.909 & 0.798 & 0.762& 0.570 & 0.649 & 0.381 & 0.448 & 0.688\\
MUSIQ (ICCV 2021) & 0.753 & 0.680 & 0.917 & {0.921} & 0.813 & 0.789 & 0.564 & 0.675 & 0.349 & 0.429 & 0.689 \\
CLIP-IQA+ (AAAI 2023) & 0.753 & 0.777 & 0.881 & 0.883 & 0.719 & 0.755 & 0.577 & 0.614 & 0.633 & 0.638 & 0.723 \\
LIQE (CVPR 2023)& {0.826} & {0.847} & {0.922} & 0.919 & 0.805 & {0.866} & 0.672 & 0.722 & 0.639 & 0.627 & 0.785\\
Q-Align (ICML 2024) & \underline{0.848} & \underline{0.879} & \textbf{0.930} & \textbf{0.933} &   \underline{0.865} & \underline{0.873} & {0.723} & {0.786} & \underline{0.743} & \textbf{0.740} & \underline{0.832}\\ \hdashline
\textbf{\textsc{\textbf{\textsc{Q-SiT-mini}}}} (\textit{Ours}) & 0.844 & 0.828 & 0.923 & 0.911 & 0.829 & 0.857 & \underline{0.775} & \underline{0.813} & 0.704 & 0.650 & 0.813 \\
\textsc{\textbf{\textbf{\textsc{Q-SiT}}}} (\textit{Ours}) & \textbf{0.888} & \textbf{0.898} & \underline{0.926} & \underline{0.924} & \textbf{0.879} & \textbf{0.897} & \textbf{0.796} & \textbf{0.854} & \textbf{0.752} & \underline{0.715} & \textbf{0.856}\\
\hline
\end{tabular}}
\resizebox{\linewidth}{!}{\begin{tabular}{l|ccccccccccc}
\hline
\makebox[0.3\textwidth][l]{\textit{Training Set:}  \textbf{SPAQ+KonIQ}\textit{$\to$Testing Set:}} & \multicolumn{2}{c}{\textbf{KonIQ}} & \multicolumn{2}{c}{\textbf{SPAQ}} &\multicolumn{2}{c}{\textbf{LIVEC}$^\textsc{Cross}$} &\multicolumn{2}{c}{\textbf{AGIQA-3K}$^\textsc{Cross}$} &\multicolumn{2}{c}{\textbf{KADID-10k}$^\textsc{Cross}$} &\multirow{2}{*}{\textbf{Avg.}} \\ \cline{1-11}
\textbf{Method}  &SRCC&PLCC & SRCC&PLCC & SRCC&PLCC & SRCC&PLCC & SRCC&PLCC \\
\hline
\textit{Zeroshot Backbone of} Q-Align  &  0.552 & 0.489 & 0.729 & 0.625 & 0.572 & 0.566 & 0.526 & 0.538 & 0.648 & 0.616 & 0.586 \\
\textit{Zeroshot Backbone of} \textbf{\textsc{\textbf{\textsc{Q-SiT-mini}}}}  & 0.728 & 0.770 & 0.860 & 0.865 & 0.758 & 0.778 & 0.733 & 0.789 & 0.503 & 0.461 & 0.724\\
\textit{Zeroshot Backbone of} \textbf{\textsc{Q-SiT}} & 0.752 & 0.842 & 0.874 & 0.869 & 0.807 & 0.808 & 0.768 & 0.786 & 0.654 & 0.655 & 0.781 \\ \hdashline
Q-Align (ICML 2024) &  \textbf{0.940} & \textbf{0.943} & \underline{0.931} & \underline{0.933} & \underline{0.879} & \underline{0.883} & 0.727 & 0.795 & 0.708 & \underline{0.692} & \underline{0.843}\\
\textbf{\textsc{\textbf{\textsc{Q-SiT-mini}}}} (\textit{Ours}) & 0.914 & 0.882  & 0.926 & 0.912 & 0.859 & 0.870 & \underline{0.773} & \underline{0.816} & \underline{0.756} & 0.621 & 0.834 \\
\textbf{\textsc{Q-SiT}} (\textit{Ours}) & \underline{0.924} & \underline{0.925} & \textbf{0.933} & \textbf{0.934} & \textbf{0.896} & \textbf{0.897} & \textbf{0.793} & \textbf{0.851} & \textbf{0.765} & \textbf{0.699} & \textbf{0.861} \\
\hline
\hline
\end{tabular}}
\label{tab:iqa}
%\vspace{-12pt}
\end{table*}

\subsubsection{Predicting the Coarse-Grained Optimal Data Mix Ratio}
Motivated by the small-to-large data-mixture transfer explored in RegMix \cite{liu2024regmix}, the performance of smaller language models under different data mix configurations can serve as a prior reference for larger language models, which allows us to approximate an optimal balance configuration with much fewer computational resources. Therefore, we first use a small-scale LMM, LLaVA-Onevision-0.5B \cite{li2024llavaov}, to efficiently assess the effectiveness of different data proportions of various datasets. Since LMMs inherently excel at interpreting, we focus on determining the optimal coarse-grained ratio between \textbf{D2} and \textbf{D3}.
We first utilize the entire \textbf{D2} dataset and systematically adjust its proportion relative to \textbf{D3} to achieve a mix ratio ranging from \textbf{D2}:\textbf{D3} = 1:0.1 to 1:1. Next, we use the entire \textbf{D3} dataset and systematically adjust its proportion relative to \textbf{D2} to achieve a mix ratio ranging from \textbf{D2}:\textbf{D3} = 1:1 to 1:10. We conduct training using these mixed datasets and evaluate model performance on a validation set, recording the mean performance scores for both image quality scoring and image interpreting.
To determine the optimal ratio, we fit a regression model that predicts the best balance between \textbf{D2} and \textbf{D3}:
\begin{equation}
r^{D2:D3} = \arg\max_{r_{A}} \mathbb{A}(D2_{r_A}, D3_{1-r_A}),
\end{equation} 
where $r^{D2:D3}$ represents the optimal mix ratio, and $\mathbb{A}(\cdot)$ denotes the fitted regression model.
Next, we incorporate \textbf{D1} into the mixed dataset \textbf{$|$D2$^*$+D3$^*$$|$} (controlled by $r^{D2:D3}$) and reassess model performance with various mix ratios. A second regression model $\mathbb{B}(\cdot)$ is then employed to predict the optimal proportion between \textbf{$|$D2$^*$+D3$^*$$|$} and \textbf{D1}:
\begin{equation}
r^{|D2^*+D3^*|:D1} = \arg\max_{r_B} \mathbb{B}(|D2^*+D3^*|_{r_B}, D1_{1-r_B}),
\end{equation}
where the mix ratio of \textbf{D1}:\textbf{D2}:\textbf{D3} (defined by $r^{D2:D3}$ and $r^{|D2^*+D3^*|:D1}$) is recorded as the \textit{Coarse-Grained Optimal Data Mix Ratio}.
At this stage, we also track the \textit{validation loss ratio} for image quality scoring and interpreting tasks, serving as the \textit{reference loss ratio}:
\begin{equation}
\lambda_{\mathcal{L}} = \frac{\mathcal{L}_{\mathcal{S}}}{\mathcal{L}_{\mathcal{I}}},
\end{equation}
where $\lambda^{loss}$ represents the \textit{reference loss ratio}, and $\mathcal{L}_{\mathcal{S}}$ and $\mathcal{L}_{\mathcal{I}}$ denote the validation loss for image quality scoring and interpreting tasks, respectively.

\subsubsection{Adjusting the Fine-Grained Optimal Data Mix Ratio}
Using the \textit{Coarse-Grained Optimal Data Mix Ratio}, we train our primary model (LLaVA-Onevision-7B). At the end of each training epoch, we monitor the validation loss for both tasks. If the loss proportion for a specific task deviates significantly from the \textit{reference loss ratio} $\lambda_{\mathcal{L}}$, we infer that additional training data is required based on the data mixture principles \cite{ye2024datamix}.
To dynamically adjust the dataset proportions, we increase the corresponding dataset size in the next epoch until the validation loss ratio aligns with the \textit{reference loss ratio}:
\begin{align}
& \text{If } \frac{\mathcal{L'}^{t}_{\mathcal{S}}}{\mathcal{L'}^{t}_{\mathcal{I}}}   < \lambda_{\mathcal{L}}, \quad \uparrow |D2^*\!\!+\!\!D3^*|, \\
& \text{If } \frac{\mathcal{L'}^{t}_{\mathcal{S}}}{\mathcal{L'}^{t}_{\mathcal{I}}} > \lambda_{\mathcal{L}}, \quad \uparrow |D1|,
\end{align}
where $\mathcal{L'}^{t}_{\mathcal{S}}$ and $\mathcal{L'}^{t}_{\mathcal{I}}$ denote the validation loss of $t$-th training epoch for the primary model, $\uparrow$ represents the data increase operating.
Through this fine-grained adjustment strategy, we ensure efficient task balancing, optimizing both image quality assessment and interpreting performance.

\begin{figure}[!h]
    \centering
    \includegraphics[width=\linewidth]{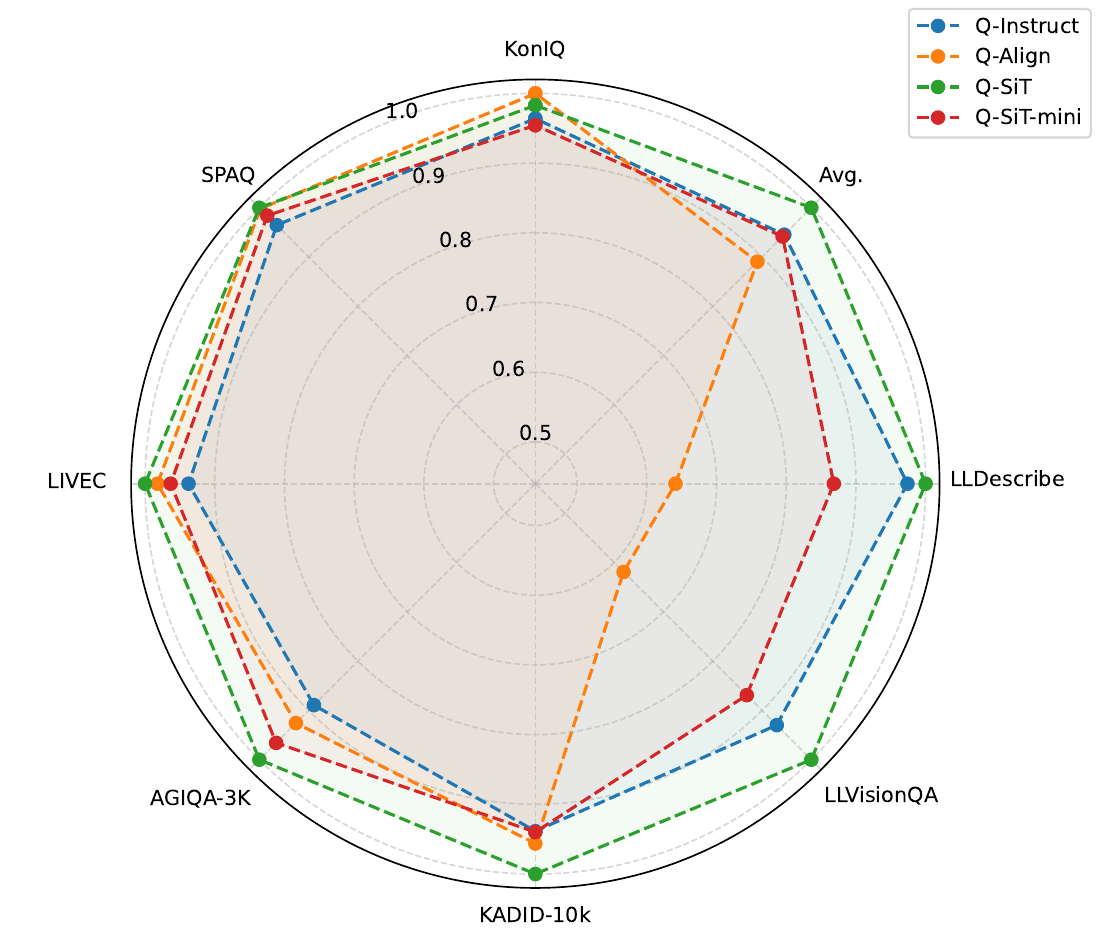}
    \caption{An overview of the performance across a series of image quality LMMs. Q-Instruct~\cite{q-instruct} is trained only on image quality interpreting data, while Q-Align~\cite{qalign} is trained only on image quality scoring data. \textbf{\textsc{Q-SiT}} and \textbf{\textsc{Q-SiT-mini}} are the proposed models, which are trained on both scoring and interpreting data. Performance on IQA datasets is measured as \textit{(SRCC+PLCC)/2}. The specific scale is determined by normalizing each performance value against the maximum performance value observed for the corresponding dataset. \textbf{`Avg.'} represents the model's average performance across all datasets.}
    \label{fig:radar}
    %\vspace{-12pt}
\end{figure}

\section{Experiment}

\subsection{Experiment Settings}

In our experiments, we set the batch size to 64 and the learning rate to 1e-5. We use LLaVA-Onevision-0.5B and LLaVA-Onevision-7B as the small and primary LMM models for \textbf{\textsc{Q-SiT}}, respectively. For training the small LMM, we stop once the validation loss plateaus, indicating no further improvement. For the primary model, we train for a default of 3 epochs. Training is performed on 8 NVIDIA A800 80G GPUs, and inference latency is reported on a single NVIDIA A800 80G GPU.
The employed datasets are described as follows:
\begin{itemize}
    \item For the \textbf{D1} data, we select KonIQ \cite{hosu2020koniq} and SPAQ \cite{spaq}. 
    \item For the \textbf{D2} data, we use the constructed image quality interpreting data discussed in Sec.\ref{sec:qins}.
    \item For the \textbf{D3} data, we use the LLaVA-150K \cite{llava} dataset.
    \item For \textbf{Validation}, we use 10\% of the training data from KonIQ and SPAQ for the image quality scoring task, and for the image quality interpreting task, we select {LLVisionQA}-{\textit{dev}} \cite{wu2024qbench} as the validation set.
    \item For \textbf{Testing}, we choose KonIQ (\textit{testing sets}), SPAQ (\textit{testing sets}), LIVEC~\cite{livechallenge}, AGIQA-3K~\cite{agiqa3k}, and KADID-10k~\cite{kadid} as scoring test sets.  We choose {LLVisionQA}-{\textit{test}} and LLDescribe \cite{wu2024qbench} as interpreting test sets. 
\end{itemize}
It is important to note that all datasets are cleaned to ensure no overlap between the training/validation and testing sets.

% \begin{figure*}[t]\small
%     \centering
%     \subfloat[]{
%         \includegraphics[width=0.48\linewidth]{figs/d2_d3_ratio.pdf}
%         \label{fig:d2_d3_ratio}
%     }
%     \hfill
%     \subfloat[]{
%         \includegraphics[width=0.48\linewidth]{figs/d1_d2d3_ratio.pdf}
%         \label{fig:d1_d2d3_ratio}
%     }
%     \caption{Visualization of the fitted curves for predicting the coarse-grained optimal data mix ratio. A polynomial regression model with a default degree of 4 is employed for the fitting process.}
%     \label{fig:data_mix_ratios}
% \end{figure*}

\begin{table*}[!htp]\small
    \centering
    \renewcommand\arraystretch{1.05}
    \renewcommand\tabcolsep{10pt}
    \caption{Results on the {\tt test} subset of \textbf{LLVisionQA} for the image quality \textbf{Perception Interpreting} ability of LMMs. Best in \textbf{bold} and second \underline{underlined}. The {Q-Instruct*} is retrained with the same LMM backbone (\textit{LLaVA-Onevision-7B}) of Q-SiT.}
    \vspace{-8pt}
    \resizebox{\linewidth}{!}{\begin{tabular}{l|ccc|cc|cc|c}
    \hline
    \hline
        \textbf{Sub-categories} & \multicolumn{3}{c|}{\textbf{Question Types}} & \multicolumn{4}{c|}{\textbf{Quadrants of Low-level Concerns}} & \multirow{3}{*}{{\textit{Overall$\uparrow$}}} \\ \cline{1-8}
        \multirow{2}{*}{\textbf{Model} \textit{(variant)}}  & \multirow{2}{*}{\textit{Yes-or-No$\uparrow$}}& \multirow{2}{*}{\textit{What$\uparrow$}} & \multirow{2}{*}{\textit{How$\uparrow$}} & \multirow{2}{*}{\textit{Distortion$\uparrow$}} & \multirow{2}{*}{\textit{Other$\uparrow$}} & \textit{In-context}  &\textit{In-context}  \\
        &&&&&&\textit{Distortion$\uparrow$}& \textit{Other$\uparrow$} \\ \hline
    \textit{\textbf{Test Set /}}\textit{ random guess} & 50.00\% & 28.48\% & 33.30\% & 37.24\% & 38.50\% & 39.13\% & 37.10\% & 37.94\% \\ \cline{1-9}
    \multicolumn{9}{l}{\textit{{Performance of Small LMMs}}} \\ \hline
         InternVL2.5 \textit{(QwenLM2-0.5B)}~\cite{internvl2.5} & \underline{72.55\%} & \underline{65.89\%} & 52.81\% & 57.32\% & \textbf{67.53\%} & 57.73\% & \underline{72.39\%} & \underline{63.31\%}\\
         mPLUG-Owl3  \textit{(QwenLM2-0.5B)}~\cite{ye2024mplugowl3} & 50.43\% & 59.56\% & 53.07\% & 46.15\% & 60.16\% & 50.39\% & 60.67\% & 53.42\% \\
         LLaVA-Onevision \textit{(QwenLM2-0.5B)}~\cite{li2024llavaov} & 66.77\% & \textbf{67.72\%} & \textbf{58.12\%} & \underline{60.25\%} & 61.48\% & \underline{57.75\%} & \textbf{77.37\%} & 62.49\%\\
         \textbf{\textsc{Q-SiT-mini}} \textit{(QwenLM2-0.5B)} (\textit{Ours}) & \textbf{75.81\%} & 65.70\% & \underline{54.56\%} & \textbf{70.03\%} & \underline{61.80\%} & \textbf{61.18\%} & 69.38\% & \textbf{64.88\%} \\ \hline
         \multicolumn{9}{l}{\textit{{Performance of Around 7B Param. LMMs}}} \\ \hline
         Qwen2.5-VL \textit{(QwenLM2.5-7B)}~\cite{Qwen2.5-VL} & \textbf{80.57\%} & \underline{81.59\%} & \underline{66.81\%} & 71.85\% & 75.45\% & \textbf{74.35\%} & 83.65\% & \underline{75.22\%} \\
         InternVL2.5 \textit{(InternLM2.5-7B)}~\cite{internvl2.5} & 78.84\% & 79.67\% & 66.13\% & 69.23\% & \textbf{75.98\%} & 69.18\% & \textbf{85.15\%} & 73.80\%   \\
         mPLUG-Owl3  \textit{(QwenLM2-7B)}~\cite{ye2024mplugowl3} & 78.72\% & 79.77\% & \textbf{67.45\%} & \underline{73.44\%} & 71.74\% & \underline{71.19\%} & \underline{84.89\%} & 74.21\% \\
         InfiMM (\textit{Zephyr-7B})~\cite{InfiMM} & 61.31\% & 56.61\% & 49.58\% & 47.79\% & 62.05\% & 51.71\% & 67.68\% & 56.05\%\\
         Fuyu-8B (\textit{Persimmon-8B})~\cite{fuyu-8b} & 62.22\% & 35.79\% & 36.62\% & 41.07\% & 49.40\% & 45.89\% & 49.04\% & 45.75\% \\
         BakLLava (\textit{Mistral-7B})~\cite{bakllava} & 66.46\% & 61.48\% & 54.83\% & 51.33\% & 63.76\% & 56.52\% & {78.16}\% & 61.02\% \\
         SPHINX~\cite{sphinx} & {74.45}\% & {65.50}\% & {62.13}\% & {59.11}\% & {73.26}\% & {66.09}\% & {77.56}\% & {67.69}\% \\
         mPLUG-Owl2 \textit{(LLaMA-7B)}~\cite{mplug2} & {72.26}\% & 55.53\% & {58.64}\% & 52.59\% & {71.36}\% & 58.90\% & 73.00\% & 62.68\% \\
        LLaVA-v1.5 \textit{(Vicuna-v1.5-7B)}~\cite{improvedllava} & 64.60\% & 59.22\% & 55.76\% & 47.98\% & {67.30}\% & {58.90}\% & {73.76}\% & 60.07\% \\
        LLaVA-v1.5 \textit{(Vicuna-v1.5-13B)}~\cite{improvedllava} & 64.96\% & {64.86}\% & 54.12\% & 53.55\% & {66.59}\% & {58.90}\% & 71.48\% & 61.40\% \\
        InternLM-XComposer-VL \textit{(InternLM)}~\cite{xcomposer} & 68.43\% & {62.04}\% & {61.93}\% & {56.81}\% & {70.41}\% & 57.53\% & {77.19}\% & {64.35}\% \\
        IDEFICS-Instruct  \textit{(LLaMA-7B)}~\cite{idefics} & 60.04\% & 46.42\% & 46.71\% & 40.38\% & 59.90\% & 47.26\% & 64.77\% & 51.51\% \\
        Qwen-VL-7B \textit{(QwenLM-7B)}~\cite{Qwen-VL} & 65.33\% & {60.74}\% & {58.44}\% & 54.13\% & 66.35\% & 58.22\% & {73.00}\% & {61.67}\% \\
        Shikra (\textit{Vicuna-7B})~\cite{shikra} & 69.09\% & 47.93\% & 46.71\% & 47.31\% & 60.86\% & 53.08\% & 64.77\% & 55.32\% \\
        Otter-v1 \textit{(MPT-7B)}~\cite{otter} & 57.66\% & 39.70\% & 42.59\% & 42.12\% & 48.93\% & 47.60\% & 54.17\% & 47.22\% \\
        InstructBLIP \textit{(Flan-T5-XL)}~\cite{iblip} & {69.53}\% & 59.00\% & {56.17}\% & {57.31}\% & 65.63\% & 56.51\% & 71.21\% & {61.94}\% \\
        InstructBLIP \textit{(Vicuna-7B)}~\cite{iblip} & {70.99}\% & 51.41\% & 43.00\% & 45.00\% & 63.01\% & 57.19\% & 64.39\% & 55.85\% \\
        VisualGLM-6B \textit{(GLM-6B)}~\cite{glm} & 61.31\% & 53.58\% & 44.03\% & 48.56\% & 54.89\% & 55.48\% & 57.79\% & 53.31\% \\
        mPLUG-Owl \textit{(LLaMA-7B)}~\cite{mplugowl} & {72.45}\% & 54.88\% & 47.53\% & 49.62\% & 63.01\% & {62.67}\% & 66.67\% & 58.93\% \\
        LLaMA-Adapter-V2~\cite{llamaadapterv2} & 66.61\% & 54.66\% & 51.65\% & {56.15}\% & 61.81\% & {59.25}\% & 54.55\% & 58.06\% \\
        LLaVA-v1 (\textit{Vicuna-13B})~\cite{llava} & 57.12\% & 54.88\% & 51.85\% & 45.58\% & 58.00\% & 57.19\% & 64.77\% & 54.72\% \\
        MiniGPT-4 \textit{(Vicuna-13B)}~\cite{minigpt4} & 60.77\% & 50.33\% & 43.00\% & 45.58\% & 52.51\% & 53.42\% & 60.98\% & 51.77\% \\
        {Q-Instruct* \textit{(QwenLM2-7B)}} & {78.80\%} & {76.20\%} & {63.50\%} & {68.80\%} & {73.90\%} & {70.10\%} & {80.50\%} & {72.30\%} \\
        \textbf{\textsc{Q-SiT}} \textit{(QwenLM2-7B)} (\textit{Ours}) & \underline{79.63\%} & \textbf{83.40\%} & 64.09\% & \textbf{74.70\%} & \underline{75.69\%} & 71.05\% & 83.26\% & \textbf{75.65\%} \\ 
        \hline
        \multicolumn{9}{l}{\textit{{Performance of Proprietary LMMs}}} \\ \hline
        \textbf{Qwen-VL-Plus} (\textit{Closed-source})  & \textbf{82.28\%} & \underline{81.50\%} & 67.46\% & \textbf{77.29\%} & \underline{74.15\%} & 75.59\% & 81.31\% & \underline{76.79\%} \\
        \textbf{Gemini-1.5-Pro} (\textit{Closed-source})  & 79.18\% & 76.77\% & \textbf{70.60\%} & 73.05\% & 73.53\% & 72.93\% & \underline{82.53\%} & 74.68\% \\ 
         \textbf{GPT-4V} (\textit{Closed-source}) & 77.72\% & 78.39\% & 66.45\% & 71.01\% & 71.07\% & \textbf{79.36\%} & 78.91\% & 74.10\%  \\
         \textbf{GPT-4o} (\textit{Closed-source, 2024-0806})  & \underline{82.26\%} & \textbf{83.36\%} & \underline{70.40\%} & \underline{76.90\%} & \textbf{76.12\%} & \underline{77.29\%} & \textbf{84.35\%} & \textbf{77.88\%} \\ \hline
         \multicolumn{9}{l}{\textit{{Performance of Humans}}} \\ \hline
         \textit{Junior-level \textit{Human}} &82.48\% & 79.39\% & 60.29\% & 75.62\% & 72.08\% & 76.37\% & 73.00\% & 74.31\%  \\
        \textit{Senior-level \textit{Human}} &84.31\% & 88.94\% & 72.02\% & 79.65\% & 79.47\% & 83.90\% & 87.07\% & 81.74\%  \\ \hline \hline
   \end{tabular}}
    \label{tab:perception}
\end{table*}

\subsection{Details about Data Mix Ratio}
\noindent \textbf{Coarse-Grained Optimal Data Mix Ratio.}
We first predict the optimal mix ratio for the interpreting task using datasets \textbf{D2} and \textbf{D3}. The cleaned \textbf{D2} contains approximately 150K instruction pairs. Initially, we include the entire \textbf{D2} dataset and randomly sample [1.5K, 3K, ..., 150K] instances from \textbf{D3} for mixed training. The performance is recorded based on the results on {LLVisionQA}-{\textit{dev}}, repeating the experiment three times to mitigate randomness. Similarly, \textbf{D3} also contains around 150K instruction pairs, and we repeat the same process to record its performance.
We then fit a polynomial model based on the observed performance results and the corresponding mix ratios, determining the optimal predicted ratio of \textbf{D2}:\textbf{D3} = 2.42. Next, considering that \textbf{D1} contains approximately 16K instances, we use 16K as the base and mix $|$\textbf{D2$^*$+D3$^*$}$|$ following a ratio of [0.1, 0.2, ..., 1, 2, ..., 10]. We then evaluate the performance on both {LLVisionQA}-{\textit{dev}} and an average score across IQA datasets to determine the final performance. Using this ratio, we again fit a polynomial model, obtaining the predicted optimal ratio of $|$\textbf{D2$^*$+D3$^*$}$|$:\textbf{D1} = 3.54. 

This gives us the \textit{Coarse-Grained Optimal Data Mix Ratio} and the details are presented in Fig.\ref{fig:data_mix_ratios}.
It is worth noting that, to reduce training time and computational costs, these experiments are conducted on LLaVA-OneVision-0.5B. Additionally, we record the validation loss ratio between image quality scoring and interpreting tasks, which is approximately 1:4.66.

\noindent \textbf{Fine-Grained Optimal Data Mix Ratio.} Based on the coarse-grained ratio, we determine an approximate mix of D1:D2:D3 = 1.00: 2.50: 1.04. Using this ratio, we train the primary \textbf{\textsc{Q-SiT}} model based on LLaVA-OneVision-7B. After each training iteration, we compute the corresponding validation loss for both image quality scoring and interpreting tasks and dynamically adjust the data mix ratio as described in Algorithm~\ref{al:mix}. This process continues until the primary model training is complete (\textit{usually no more than 3 epochs}).

\begin{table*}\small
    \centering
    \renewcommand\arraystretch{1.1}
    \renewcommand\tabcolsep{4.4pt}
        \caption{Results on the image quality \textbf{Description Interpreting} ability of LMMs. $P_i$ denotes frequency for score $i$. The {Q-Instruct*} is retrained with the same LMM backbone (\textit{LLaVA-Onevision-7B}) of Q-SiT.}
        \vspace{-8pt}
    \resizebox{\linewidth}{!}{\begin{tabular}{l|cccc|cccc|cccc|c}
    \hline
    \hline
        \textbf{Dimensions} & \multicolumn{4}{c|}{\textbf{Completeness}} & \multicolumn{4}{c|}{\textbf{Precision}} & \multicolumn{4}{c|}{\textbf{Relevance}} & \multirow{2}{*}{\textit{Sum.$\uparrow$}} \\ \cdashline{1-13}
        \textbf{Model} (\textit{variant}) & $P_0$ & $P_1$ & $P_2$ & \textit{score$\uparrow$}   &  $P_0$ & $P_1$ & $P_2$ & \textit{score$\uparrow$}   & $P_0$ & $P_1$ & $P_2$  & \textit{score$\uparrow$} \\ \hline
         \multicolumn{14}{l}{\textit{{Performance of Small LMMs}}} \\ \hdashline
         InternVL2.5 \textit{(QwenLM2-0.5B)}~\cite{internvl2.5} & 22.27\% & 67.28\% & 10.19\% & \underline{0.88} & 35.07\% & 25.30\% & 39.63\% & 1.05 & 17.97\% & 58.58\% & 23.45\% & 1.05 & 2.98\\
         mPLUG-Owl3  \textit{(QwenLM2-0.5B)}~\cite{ye2024mplugowl3} & 26.88\% & 52.92\% & 19.25\% & \textbf{0.91} & 31.80\% & 25.25\% & 42.88\% & \textbf{1.11} & 8.54\% & 60.76\% & 30.70\% & 1.22 & \underline{3.25} \\
         \textbf{\textsc{\textbf{\textsc{Q-SiT-mini}}}} \textit{(QwenLM2-0.5B)} (\textit{Ours}) & 27.61\% & 64.01\% & 8.38\% & 0.81 & 29.09\% & 32.37\% & 38.41\% & \underline{1.09} & 3.83\% & 21.10\% & 74.59\% & \textbf{1.70} & \textbf{3.60}\\ \hline
         \multicolumn{14}{l}{\textit{{Performance of Around 7B Param. LMMs}}} \\ \hdashline
         Qwen2.5-VL \textit{(QwenLM2.5-7B) }~\cite{Qwen2.5-VL} & 24.57\% & 50.57\% & 24.24\% & 0.99 & 52.55\% & 22.95\% & 24.47\% & 0.72 & 0.06\% & 35.19\% & 64.4\% & 1.64 & 3.35 \\
         InternVL2.5 \textit{(InternLM2.5-7B)}~\cite{internvl2.5} & 23.56\% & 50.12\% & 26.32\% & 1.03 & 23.87\% & 28.21\% & 47.92\% & 1.24 & 1.38\% & 11.59\% & 86.68\% & \underline{1.85} & 4.12 \\
         mPLUG-Owl3  \textit{(QwenLM2-7B)}~\cite{ye2024mplugowl3} & 27.54\% & 37.08\% & 35.13\% & 1.07 & 28.93\% & 20.46\% & 50.13\% & 1.21 & 8.72\% & 48.64\% & 42.11\% & 1.33 & 3.61 \\
         LLaVA-Onevision \textit{(QwenLM2-7B)}~\cite{li2024llavaov} & 8.10\% & 72.24\% & 20.08\% & \underline{1.11} & 29.67\% & 33.79\% & 36.44\% & 1.07 & 0.46\% & 13.22\% & 86.27\% & \underline{1.86} & 4.05 \\
        InfiMM (\textit{Zephyr-7B})~\cite{InfiMM} & 29.61\% & 62.32\% & 7.77\% & 0.77 & 29.25\% & 31.90\% & 38.51\% & 1.08 & 2.16\% & 22.72\% & 74.58\% & 1.71 & 3.58 \\
        Fuyu-8B (\textit{Persimmon-8B})~\cite{fuyu-8b} & 25.54\% & 61.00\% & 13.46\% & 0.88 & {41.96}\% & {32.76}\% & 25.28\% & 0.83 & 2.99\% & 11.34\% & 85.67\% & 1.82 & 3.53 \\
        BakLLava (\textit{Mistral-7B})~\cite{bakllava} &24.31\% & 51.22\% & 24.47\% & 1.00 & {49.23}\% & 24.11\% & 26.66\% & 0.77 & 1.25\% & 36.22\% & {62.53}\% & {1.61} & 3.38 \\
        SPHINX~\cite{sphinx} & 27.96\% & 64.36\% & 7.33\% & 0.79 & 26.16\% & 32.42\% & 41.01\% & 1.14 & 1.69\% & 23.00\% & 74.61\% & 1.72 & 3.65 \\
        mPLUG-Owl2 (\textit{LLaMA-7B})~\cite{mplug2} & 27.71\% & 38.58\% & 33.71\% & 1.06 &28.11\% & 19.78\% & 52.11\% & 1.24 &7.91\% & 48.18\% & 43.91\% & 1.36 &  3.67 \\
        LLaVA-v1.5 (\textit{Vicuna-v1.5-7B})~\cite{improvedllava} & 27.48\% & 54.74\% & 17.78\% & 0.90 & 30.51\% & 26.04\% & 43.45\%  & 1.13 &  10.85\% & 60.34\% & 28.81\% & 1.18 & 3.21 \\
        LLaVA-v1.5 (\textit{Vicuna-v1.5-13B})~\cite{improvedllava} & 27.68\% & 53.78\% & 18.55\% & 0.91 & 25.45\% & 21.47\% & 53.08\% & \textbf{1.28} & 6.31\% & 58.75\% & 34.94\% & 1.29 & 3.47 \\
        InternLM-XComposer-VL \textit{(InternLM)}~\cite{xcomposer} & 19.94\% & 51.82\% & 28.24\% & {1.08} & 22.59\% &  28.99\% & 48.42\% & \underline{1.26} & 1.05\% & 10.62\% & 88.32\% & \textbf{1.87} & \textbf{4.21} \\
        IDEFICS-Instruct \textit{(LLaMA-7B)}~\cite{idefics} & 28.91\% & 59.16\% & 11.93\% & 0.83 & 34.68\% & 27.86\% & 37.46\% & 1.03 & 3.90\% & 59.66\% & 36.44\% & 1.33 & 3.18 \\
        Qwen-VL \textit{(QwenLM)}~\cite{Qwen-VL} & 26.34\% & 49.13\% & 24.53\% & 0.98 & {50.62}\% & 23.44\% & 25.94\% & 0.75 & 0.73\% & 35.56\% & {63.72}\% & {1.63} & 3.36 \\
        Shikra (\textit{Vicuna-7B})~\cite{shikra} & 21.14\% & {68.33}\% & 10.52\% & 0.89 & 30.33\% & 28.30\% & 41.37\% & 1.11 & 1.14\% & {64.36}\% & 34.50\% & 1.33 & 3.34 \\
        Otter-v1 \textit{(MPT-7B)}~\cite{otter} & 22.38\% & 59.36\% & 18.25\% & 0.96 & {40.68}\% & {35.99}\% & 23.33\% & 0.83 & 1.95\% & 13.20\% & {84.85}\% & {1.83} & 3.61 \\
        Kosmos-2~\cite{kosmos2} & 8.76\% & {70.91}\% & 20.33\% & \textbf{1.12} & 29.45\% & {34.75}\% & 35.81\% & 1.06 & 0.16\% & 14.77\% & {85.06}\% & {1.85} & {4.03} \\
        InstructBLIP \textit{(Flan-T5-XL)}~\cite{iblip} & 23.16\% & 66.44\% & 10.40\% & 0.87 & 34.85\% & 26.03\% & 39.12\% & 1.04 & {14.71}\% & 59.87\% & 25.42\% & 1.11 & 3.02 \\
        InstructBLIP \textit{(Vicuna-7B)}~\cite{iblip} & 29.73\% & 61.47\% & 8.80\% & 0.79 & 27.84\% & 23.52\% & 48.65\% & 1.21 & {27.40}\% & 61.29\% & 11.31\% & 0.84 & 2.84 \\
        VisualGLM-6B \textit{(GLM-6B)}~\cite{glm} & {30.75}\% & 56.64\% & 12.61\% & 0.82 & {38.64}\% & 26.18\% & 35.18\% & 0.97 & 6.14\% & {67.15}\% & 26.71\% & 1.21 & 2.99 \\
        mPLUG-Owl \textit{(LLaMA-7B)}~\cite{mplugowl} & 28.28\% & 37.69\% & {34.03}\% & {1.06} & 26.75\% & 18.18\% & {55.07}\% & \textbf{1.28} & 3.03\% & 33.82\% & 63.15\% & 1.60 & {3.94} \\
        LLaMA-Adapter-V2~\cite{llamaadapterv2} & 30.44\% & 53.99\% & 15.57\% & 0.85 & 29.41\% & 25.79\% & 44.80\% & 1.15 & 1.50\% & 52.75\% & 45.75\% & 1.44 & 3.45 \\
        LLaVA-v1 (\textit{Vicuna-13B})~\cite{llava} & {34.10}\% & 40.52\% & {25.39}\% & 0.91 & 30.02\% & 15.15\% & {54.83}\% & 1.25 & 1.06\% & 38.03\% & 60.91\% & 1.60 & {3.76} \\
        MiniGPT-4 (\textit{Vicuna-13B})~\cite{minigpt4} & {34.01}\% & 32.15\% & {33.85}\% & {1.00} & 29.20\% & 15.27\% & {55.53}\% & \underline{1.26} & 6.88\% & 45.65\% & 47.48\% & 1.41 & 3.67 \\
        {Q-Instruct* \textit{(QwenLM2-7B)}} & {15.68\%} & {58.32\%} & {26.00\%} & {1.10} & {21.98\%} & {34.01\%} & {44.01\%} & {1.22} & {1.21\%} & {20.90\%} & {77.90\%} & {1.77} & {4.09} \\
        \textbf{\textsc{Q-SiT}} \textit{(QwenLM2-7B)} (\textit{Ours}) & 18.35\% & 52.43\% & 29.21\% & \underline{1.11} & 23.47\% & 30.02\% & 46.49\% & 1.23 & 3.77\% & 9.62\% & 86.60\% & 1.83 & \underline{4.17}\\
    \hline
    \hline
    \end{tabular}}
    \vspace{-10pt} 
    \label{tab:description}
\end{table*}

\begin{figure}[t]
    \centering
    \includegraphics[width=\linewidth]{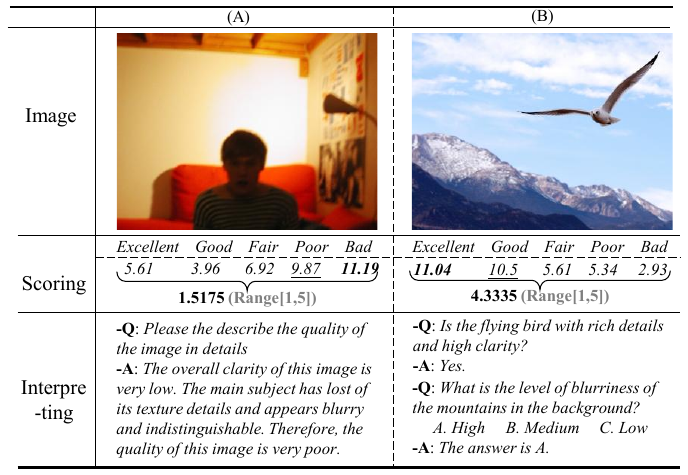}
    \caption{Visualization results of the \textbf{\textsc{Q-SiT}} on the image (A) with low quality and the image (B) with high image quality respectively. }
    \label{fig:visualization}
\end{figure}

\subsection{Quick Look of the Performance}
To quickly demonstrate the performance of \textbf{\textsc{Q-SiT}}, we provide a concise summary in Fig.~\ref{fig:radar}. From this, we can draw two interesting conclusions:
\textbf{1}) Q-Instruct exhibits a relatively balanced performance but does not stand out in IQA tasks. This suggests that training solely on image quality interpreting data can still endow LMMs with a fundamental ability for image quality scoring.
\textbf{2}) In contrast, Q-Align shows a highly unbalanced performance, it performs well in IQA but significantly deteriorates in tasks related to image quality interpreting. This indicates that training exclusively on image quality scoring data does not equip the model with a proper ability to interpret image quality. The main reason for this is that the question-answer format in scoring tasks is relatively fixed, which severely impacts the performance of interpreting tasks, where a more flexible Q\&A format is required.
Our proposed \textbf{\textsc{Q-SiT}}, on the other hand, delivers highly competitive and well-balanced performance across all tasks.

\subsection{Image Quality Scoring Performance}

For image quality scoring performance, we first evaluate the conventional setting where models are trained on a single dataset. As shown in Table~\ref{tab:iqa}, while CLIP-based methods (CLIP-IQA+ and LIQE) achieve only comparable or even worse performance in intra-dataset settings compared to the visual-only state-of-the-art, MUSIQ. The proposed \textbf{\textsc{Q-SiT}} slightly underperforms Q-Align in intra-dataset evaluations due to minor perturbations in image quality interpreting data. However, thanks to the injection of general quality knowledge from image quality interpreting data, \textbf{\textsc{Q-SiT}} demonstrates significant improvements in cross-dataset (OOD generalization) settings, outperforming visual-only methods by over \textbf{10\%}, and surpassing CLIP-IQA+ and LIQE by \textbf{8\%} and \textbf{6\%}, respectively. Furthermore, \textbf{\textsc{Q-SiT}} significantly outperforms Q-Align in cross-dataset settings, indicating its enhanced robustness. In summary, under the same data conditions, the LMM-based \textbf{\textsc{Q-SiT}} shows highly competitive performance.

We further evaluate the mixed-dataset scenario for \textbf{\textsc{Q-SiT}} on IQA, as shown in Table~\ref{tab:iqa}, to find out whether it can retain or even improve accuracy on individual datasets when datasets are combined via simple concatenation (\textit{SPAQ+KonIQ}). With general observation, the \textbf{\textsc{Q-SiT}} achieves the best performance across nearly all datasets except KonIQ. Moreover, after training on the mixed dataset, \textbf{\textsc{Q-SiT}} achieves further improvements across all IQA datasets, including all cross-dataset settings. This suggests that mixing datasets helps expand the scope of IQA knowledge, enabling the model to better predict image quality. Notably, while \textbf{\textsc{Q-SiT-mini}} does not perform as strongly as \textbf{\textsc{Q-SiT}} or Q-Align, it still significantly outperforms other baseline models and demonstrates promising cross-dataset generalization, paving the way for the development of lightweight LMM-based image quality evaluators.

\subsection{Image Quality Interpreting Performance}

\noindent \textbf{LLVisionQA Performance.}
The performance of LMMs on the LLVisionQA dataset (\textit{measuring the image quality question-answering ability on low-level attributes}) is shown in Table~\ref{tab:perception}, from which we can draw several key conclusions:
\textbf{1}) Generally speaking, as the parameter size of LMMs increases, their ability to interpret image quality improves significantly. Notably, some closed-source LMMs have already surpassed Junior-level human performance and are within \textbf{4}\% of Senior-level human performance.
\textbf{2}) \textbf{\textsc{Q-SiT}} and \textbf{\textsc{Q-SiT-mini}} achieve the best performance within their respective parameter groups (\textit{0.5B \& 7B}), indicating that our proposed \textbf{\textsc{Q-SiT}} model demonstrates strong image quality interpreting capabilities across different model scales.
\textbf{3}) Specifically, \textbf{\textsc{Q-SiT}} and \textbf{\textsc{Q-SiT-mini}} excel in \textit{Yes-or-No} and \textit{What} questions, further validating their effectiveness in image quality assessment.
\textbf{4}) Regarding low-level concerns, \textbf{\textsc{Q-SiT}} and \textbf{\textsc{Q-SiT-mini}} show a significant lead in the distortion category, likely due to their exposure to more comprehensive quality-related knowledge, enabling more precise distortion assessments.

\begin{table}[t]\small
    \centering
    \renewcommand\arraystretch{1.1}
    \renewcommand\tabcolsep{8pt}
    \caption{Ablation study results for \textbf{\textsc{Q-SiT}}. CGR and FGR denote the coarse-grained and fine-grained optimal data mix ratio strategies respectively. \textbf{\textsc{Scoring}} performance is reported as the average of SRCC and PLCC on IQA datasets, while \textbf{\textsc{Interpreting}} performance is evaluated on the LLVisionQA dataset. }
    \resizebox{\linewidth}{!}{\begin{tabular}{c|ccc|cc}
    \hline
    \hline
      \textbf{\textsc{Q-SiT}}  &  \textbf{D1} & \textbf{D2} & \textbf{D3} & \textbf{\textsc{Scoring}} & \textbf{\textsc{Interpreting}}\\
      \hline
        - & $\times$ & \checkmark & \checkmark & 0.800 & \underline{0.754}\\
        - & \checkmark & $\times$ & \checkmark & 0.857 & 0.721\\
        - & \checkmark & \checkmark & $\times$ & \underline{0.859} & 0.744\\
        \hdashline
        \textit{w/o} CGR \textit{w/o} FGR & \checkmark & \checkmark & \checkmark & 0.855 & 0.732 \\
        \textit{w} CGR \textit{w/o} FGR & \checkmark & \checkmark & \checkmark & \underline{0.859} & 0.752\\
        \textit{w} CGR \textit{w} FGR & \checkmark & \checkmark & \checkmark & \textbf{0.861} & \textbf{0.756} \\
    \hline
    \hline
    \end{tabular}}
    \label{tab:ablation}
    \vspace{-12pt}
\end{table}

\noindent \textbf{LLDescribe Performance.}
The performance of LMMs on the LLDescribe dataset (\textit{measuring the image quality describing ability of single images}) is presented in Table~\ref{tab:description}. We observe the following:
\textbf{1}) Smaller LMMs, constrained by the capabilities of their language model branch, exhibit widespread degradation in description tasks. However, \textbf{\textsc{Q-SiT-mini}} still achieves competitive performance on quality description, comparable to mid-tier 7B LMMs. This highlights the potential of running image quality description models on lightweight platforms.
\textbf{2}) Among 7B-level LMMs, \textbf{\textsc{Q-SiT}} ranks second in overall performance, demonstrating its ability to generate comprehensive and accurate descriptions related to image quality.

\noindent \textbf{Scoring Supervision on Image Quality Interpreting.}
To further examine whether image-quality scoring supervision benefits image quality interpreting, we compare \textbf{\textsc{Q-SiT}} with a Q-Instruct interpreting baseline using the same large-model backbone. 
As shown in Table~\ref{tab:perception} and Table~\ref{tab:description}, \textbf{\textsc{Q-SiT}} consistently outperforms this baseline on both LLVisionQA and LLDescribe. 
These results indicate that scoring-oriented supervision provides useful low-level quality priors for both question-answering-based and description-based image quality interpreting, thereby validating the complementary relationship between image quality scoring and image quality interpreting.

\subsection{Qualitative Analysis}
In Fig.~\ref{fig:visualization}, we visualize the scoring and interpreting results of \textbf{\textsc{Q-SiT}}. It can be observed that \textbf{\textsc{Q-SiT}} accurately determines that the quality of (B) is significantly higher than that of (A) in the scoring task. Furthermore, \textbf{\textsc{Q-SiT}} not only correctly identifies the quality issues present in (A) but also effectively perceives the rich details of the flying bird in (B).

\subsection{Ablation Study}

We further conduct ablation studies to evaluate the contribution of different datasets and training strategies. The results are presented in Table \ref{tab:ablation}. \textbf{1}) It is evident that removing \textbf{D1} significantly degrades scoring performance, while excluding \textbf{D2} notably impacts interpreting performance. \textbf{2}) Additionally, the combination of \textbf{D2} and \textbf{D3} further enhances interpreting performance, reinforcing our previous hypothesis that general semantic understanding can aid in improving image quality interpreting.
\textbf{3}) Moreover, we observe that simply mixing data without any training strategy yields suboptimal results. However, as the coarse-grained and fine-grained optimal data mix ratio strategies are progressively applied, both image quality scoring and interpreting performance improve. This demonstrates the effectiveness of our proposed training strategy.

\begin{figure}
    \centering
    \includegraphics[width=0.8\linewidth]{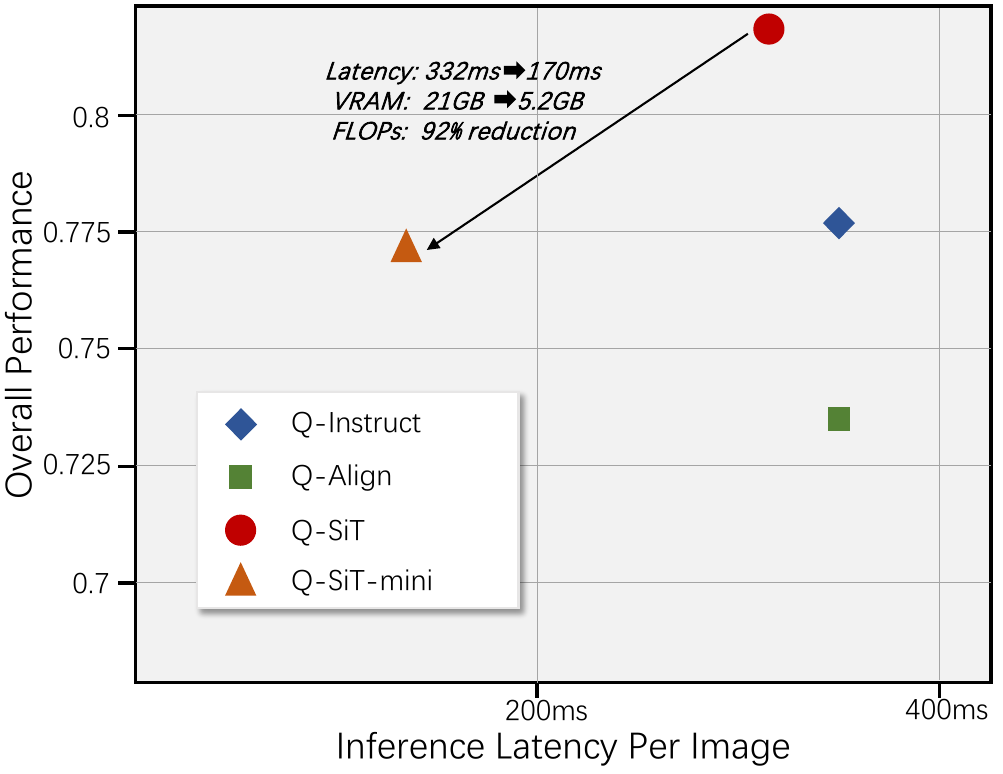}
    \caption{Computational efficiency comparison of Q-Instruct, Q-Align, \textbf{\textsc{Q-SiT}}, and \textbf{\textsc{Q-SiT-mini}} in terms of latency, VRAM usage, and FLOPs. The reported values represent the average across all test tasks, measured in a single-GPU environment using an NVIDIA A800 80G (\textit{with Transformers package}).}
    \label{fig:latency}
    \vspace{-12pt}
\end{figure}

\subsection{Computational Efficiency}
% train 0.5B 20*(1.5+7.5)/2 100h
% train 7B 10h

In this section, we focus on the computational efficiency of \textbf{\textsc{Q-SiT}} during both training and inference. To enhance efficiency, the coarse-grained data mix ratio is determined by first training a lightweight LLaVA-OneVision-0.5B model, which takes approximately 720 A800 GPU hours. Using this optimized ratio, the primary \textbf{\textsc{Q-SiT}} model, based on LLaVA-OneVision-7B, requires around 72 A800 GPU hours for training. In contrast, directly searching for the optimal mix ratio with LLaVA-OneVision-7B would demand approximately 2,800 A800 GPU hours (\textit{estimated from the experience of our training process}). Therefore, our strategy reduces computational cost by approximately \textbf{72\%} on the training end.

For inference, detailed comparison results are presented in Fig. \ref{fig:latency}. Under comparable computational costs, \textbf{\textsc{Q-SiT}} significantly outperforms Q-Instruct and Q-Align. Furthermore, compared to \textbf{\textsc{Q-SiT}}, \textbf{\textsc{Q-SiT-mini}} achieves nearly a \textbf{49\%} reduction in inference latency, requiring only \textbf{25\%} of the VRAM while reducing FLOPs by approximately \textbf{92\%} during inference. Despite these optimizations, \textbf{\textsc{Q-SiT-mini}} maintains a balanced performance in both scoring and interpreting tasks.

\section{Conclusion}

In this work, we propose \textbf{\textsc{Q-SiT}}, a unified \underline{Q}uality \underline{S}coring and \underline{I}nterpreting joint \underline{T}eaching framework, enabling large multimodal models (LMMs) to simultaneously perform image quality scoring and interpreting. Unlike prior works that treat these two tasks separately, we recognize their conceptual connection based on image-quality-specific psychophysical studies and design a framework that effectively bridges them.
To achieve this, we convert traditional IQA datasets into a learnable QA format and incorporate human-annotated interpreting data to enhance LMMs' understanding of low-level visual attributes. Furthermore, we introduce a scoring \& interpreting balance strategy, which dynamically adjusts data proportions to mitigate task interference and leverage cross-task knowledge. This ensures that the model maintains strong performance across both tasks without compromising either.
Additionally, we propose \textbf{\textsc{Q-SiT-mini}}, a lightweight 0.5B-level model optimized for efficiency, significantly reducing computational overhead while retaining competitive performance. Experimental results demonstrate that \textbf{\textsc{Q-SiT}} achieves state-of-the-art performance in both image quality scoring and interpreting, while \textbf{\textsc{Q-SiT-mini}} offers a cost-effective alternative for practical deployment.
Our findings highlight the feasibility and necessity of unifying quality scoring and interpreting within a single model. We believe this work provides a solid foundation for future research on integrating interpretability and numerical assessment in LMM-based IQA, paving the way for more explainable and efficient image quality evaluation systems.

\bibliographystyle{IEEEtran}
\bibliography{ref}

\end{document}